\documentclass[10pt,twocolumn,letterpaper]{article}

\usepackage[pagenumbers]{cvpr} % To force page numbers, e.g. for an arXiv version
\usepackage[normalem]{ulem}

\definecolor{cvprblue}{rgb}{0.21,0.49,0.74}
\usepackage[pagebackref,breaklinks,colorlinks,allcolors=cvprblue]{hyperref}
\usepackage{booktabs,multirow,graphicx,array,makecell}
\usepackage[table]{xcolor}
\usepackage{graphicx}
\usepackage{float}
\definecolor{ltgray}{gray}{0.94}
\renewcommand{\arraystretch}{1.06}
\newcolumntype{M}[1]{>{\centering\arraybackslash}m{#1}}

\usepackage{dsfont}
\usepackage[ruled,vlined]{algorithm2e}

\def\paperID{***} % *** Enter the Paper ID here
\def\confName{CVPR}
\def\confYear{2026}

\title{TALON: Test-time Adaptive Learning for On-the-Fly Category Discovery}

\author{Yanan Wu$^{1}$, Yuhan Yan$^{2}$, Tailai Chen$^{1,7}$, Zhixiang Chi$^{3}$, ZiZhang Wu$^{4}$, \\ Yi Jin$^{5}$, Yang Wang$^{6}$, Zhenbo Li$^{1}$\thanks{Corresponding Author} \\
$^{1}$College of Information and Electrical Engineering, China Agricultural University, China \\  %Beijing, 100083,
$^{2}$College of Science, China Agricultural University, China \\
$^{3}$Department of Electrical and Computer Engineering, University of Toronto, Canada \\ %Toronto, M5G1V7,
$^{4}$Institute of Brain-Inspired Intelligence and Artificial Intelligence, Fudan University, China \\ %Shanghai, 200433,
$^{5}$School of Computer and Information Technology, Beijing Jiaotong University, China \\ %Beijing, 100044,
$^{6}$Department of Computer Science and Software Engineering, Concordia University, Canada \\ %Montreal, H3G2J1,
$^{7}$Woof RoboT Co., Ltd., China \\
{\tt\small \{ynwu,conrain,lizb\}@cau.edu.cn, tailai\_chen2022@163.com, zhxchi@ece.utoronto.ca}, \\ {\tt\small  wuzizhang87@gmail.com, yjin@bjtu.edu.cn, yang.wang@concordia.ca}}
\begin{document}
\maketitle

\begin{abstract} 
On-the-fly category discovery (OCD) aims to recognize known categories while simultaneously discovering novel ones from an unlabeled online stream, using a model trained only on labeled data. Existing approaches freeze the feature extractor trained offline and employ a hash-based framework that quantizes features into binary codes as class prototypes. However, discovering novel categories with a fixed knowledge base is counterintuitive, as the learning potential of incoming data is entirely neglected. In addition, feature quantization introduces information loss, diminishes representational expressiveness, and amplifies intra-class variance. It often results in category explosion, where a single class is fragmented into multiple pseudo-classes. To overcome these limitations, we propose a test-time adaptation framework that enables learning through discovery. It incorporates two complementary strategies: a semantic-aware prototype update and a stable test-time encoder update. The former dynamically refines class prototypes to enhance classification, whereas the latter integrates new information directly into the parameter space. Together, these components allow the model to continuously expand its knowledge base with newly encountered samples. Furthermore, we introduce a margin-aware logit calibration in the offline stage to enlarge inter-class margins and improve intra-class compactness, thereby reserving embedding space for future class discovery. Experiments on standard OCD benchmarks demonstrate that our method substantially outperforms existing hash-based state-of-the-art approaches, yielding notable improvements in novel-class accuracy and effectively mitigating category explosion. The code is publicly available at \textcolor{blue}{https://github.com/ynanwu/TALON}.

\end{abstract}
\section{Introduction}
\label{sec:intro}

\begin{figure}[t]
    \centering
    \includegraphics[width=1\linewidth]{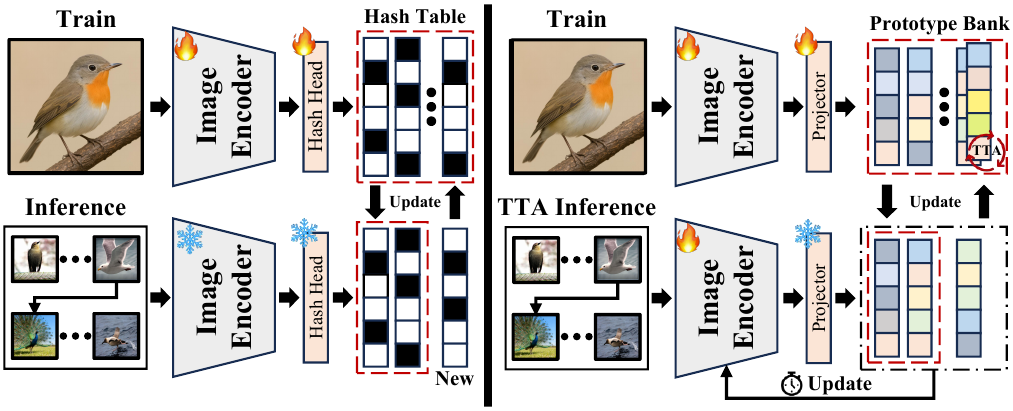}
    \caption{\textbf{At test time}, existing methods (\textit{left}) rely on \textbf{static inference} and fail to adapt to label space shifts, leading to inaccurate discovery and recognition. In contrast, our TALON (\textit{right}) \textbf{continually accumulates knowledge} from unlabeled test data to overcome this challenge. Moreover, our \textbf{hash-free framework}(\textit{right}) replaces heuristic hash-based (\textit{left}) designs, achieving higher representational expressiveness and better discovery stability.}
    \vspace{-5mm}
    \label{fig:tta}
\end{figure}

Conventional visual recognition systems \cite{he2016deep, chen2021review, hong2021spectralformer,ma2025genhancer} are typically developed under closed-world assumptions, where all categories are predefined. Such models cannot discover new concepts or generalize beyond the training set. In contrast, the human visual system recognizes unfamiliar objects by recalling prior experiences and continually expanding its conceptual memory~\cite{liu2022few}. Inspired by this capability, recent studies have explored open-world learning paradigms that enable models to identify unseen categories without explicit supervision. Among them, Novel and Generalized Category Discovery (NCD/GCD) \cite{RankStat, fini2021unified, vaze2022generalized, li2025generalized} learn to discover new classes within unlabeled data by transferring knowledge from labeled examples. However, both rely on offline training, which is impractical in dynamic environments. The continual extension of GCD (C-GCD) \cite{wu2023metagcd, roy2022class, zhao2023incremental, ma2024happy} divides learning into multiple sequential stages, where the first stage uses labeled data and later stages process large unlabeled sets. This setting still assumes that each incremental stage involves training with data covering all novel classes, which is overly idealized. Recently, On-the-Fly Category Discovery (OCD) \cite{SMILE, PHE, DiffGRE} has emerged as a more realistic formulation, consisting of an offline training phase and an online phase where the model encounters a sequential stream of unlabeled instances during test time.

Instance-based discovery and recognition, as in OCD, are inherently challenging because each instance carries limited and diverse information. To prevent optimization instability in the online stage, methods such as SMILE \cite{SMILE} and PHE \cite{PHE} freeze the feature extractor trained offline. Then, they adopt a heuristic hash-based framework, where each image feature is quantized into a binary hash code to represent class prototypes. This quantization reduces feature expressiveness and limits the information preserved in prototypes. Moreover, the binary codes are highly sensitive to intra-class variation, leading to category explosion—a phenomenon where one actual class splits into multiple pseudo-classes. To improve expressiveness, DiffGRE \cite{DiffGRE} adopts a generative framework to synthesize samples for new categories in the offline stage. However, it projects the features into much lower dimension, which is inherently ineffective. PHE \cite{PHE} attempts to enhance prototype expressiveness by assigning multiple prototypes to each class instead of one. However, all operations remain confined to binary space. Most importantly, during the online stage, none of the prototypes or encoders evolve with incoming data~\cite{chi2022metafscil}. Therefore, the model cannot truly learn from discovery.

We argue that fixing both the feature encoder and class prototypes when encountering new data is counterintuitive, as it ignores the potential to learn from incoming samples, as illustrated in Fig.~\ref{fig:tta}. To address this, we propose a test-time adaptation (TTA) framework designed explicitly for OCD, where novel classes frequently appear under semantic shift, unlike primary conventional TTA methods \cite{wang2020tent, sun2020test,wang2024distribution,chi2025plug}, which address the domain shift. Our approach updates both the model parameters and class prototypes to maximize knowledge absorption.
During the online stage, we first introduce an online decision rule to distinguish between known and novel classes. We then perform a semantic-aware prototype update that dynamically refines prototypes using a confidence-controlled exponential moving average, applying larger updates for high-confidence samples and smaller ones for uncertain cases. To further adapt the encoder in a stable manner, we employ an entropy-based objective combined with prototype-level regularization. The entropy term encourages confident predictions for incoming samples, whereas the regularization preserves alignment between features and their corresponding prototypes. This joint formulation maintains semantic consistency and ensures reliable adaptation throughout test time.

To preserve the representational expressiveness of the model, we discard heuristic hash encoding and directly operate in the continuous feature space. In the offline training stage, we further introduce a margin-aware logit calibration that enhances intra-class compactness and inter-class separability. This calibration constrains the embedding space, making it forward-compatible with emerging novel categories. Comprehensive experiments on standard benchmarks demonstrate the superiority of our approach over existing state-of-the-art methods. Moreover, the issue of category explosion is effectively mitigated, as evidenced by the significant performance gains on novel classes. 
Our main contributions are summarized as follows:
\begin{itemize}
    \item We propose a test-time adaptation framework specifically designed for on-the-fly category discovery. It jointly updates the encoder parameters and class prototypes, enabling the model to learn through discovery rather than rely on static inference.
    \item In the offline training stage, we introduce a margin-aware logit calibration that refines the embedding space by enhancing intra-class compactness and inter-class separability, ensuring forward compatibility with emerging novel categories.
    \item We demonstrate that the proposed hash-free framework effectively replaces heuristic hash-based designs, achieving higher representational expressiveness and better discovery stability.
    \item Comprehensive experiments on standard benchmarks validate the superiority of our approach over existing methods, showing significant improvements on novel classes and effective mitigation of the category explosion.
\end{itemize}

\section{Related Work}
\label{sec:related_work}

\paragraph{Category Discovery.} \textit{Novel Class Discovery} (NCD) aims to discover the novel classes within abeled data by utilizing the prior knowledge learned from labeled data~\cite{RankStat,fini2021unified,NCD, li2023modeling, zhong2021neighborhood, yang2023bootstrap}. NCD typically assumes that all unlabeled samples belong exclusively to novel classes, which rarely holds in realistic settings. Alternatively, \textit{Generalized Category Discovery} (GCD) setting ~\cite{vaze2022generalized, ma2024active, li2025generalized, wu2023metagcd, pu2023dynamic, ma2025protogcd} extends NCD by allowing unlabeled data to contain a mixture of known and novel classes. While this relaxation improves flexibility, GCD methods 
still require both known and novel categories to be jointly available during training \cite{choi2024contrastive}. This constraint leads to repetitive large-scale training when different groups of unlabeled data are continually presented to the recognition system. 
Moreover, their reliance on offline inference limits applicability in online settings where data stream continuously and instant feedback is required.

% still require known and novel classes to be jointly available during training \cite{choi2024contrastive}, leading to repetitive large-scale retraining when new unlabeled data arrive. Their reliance on offline inference also limits applicability in online settings that demand real-time feedback.

To address these limitations, Du \textit{et al.}~\cite{SMILE} introduced \textit{On-the-Fly Category Discovery} (OCD), a hash-code-based framework that removes the assumption of a predefined query set and supports instance-level feedback in streaming data settings. Building upon this idea, Zheng \textit{et al.}~\cite{PHE} proposed the prototypical hash encoding framework to mitigate the sensitivity and instability of hash-based representations. Recently, Liu \textit{et al.}~\cite{DiffGRE} argued that relying solely on labeled data is suboptimal, and developed a diffusion-based OCD framework that synthesizes novel samples to enrich offline training. Although these approaches achieve certain performance gains, they rely on fixed model parameters at test-time, limiting adaptability to the semantic shift that naturally arise in streaming data~\cite{liu2023meta}. In this work, we are the first to introduce the TTA framework into the OCD task, enabling the model to absorb new knowledge from the evolving test stream continuously. Furthermore, we propose a simple yet effective hash-free framework that enhances representation quality and overall discovery stability.

% After that  

\paragraph{Test-Time Adaptation (TTA).}  TTA aims to adapt trained models towards test data to mitigate performance degradation caused by semantic shifts~\cite{chiadapting,chilearning}. Early \cite{lim2023ttn, schneider2020improving, you2021test,wu2024test} primarily focused on improving TTA performance by 
recalibrating batch normalization statistics on test data and designing unsupervised adaptation objectives. 
TENT~\cite{wang2020tent} adapts the affine parameters of BN layers by minimizing the entropy of model predictions. MEMO~\cite{zhang2022memo} enhances adaptation by augmenting each test sample and minimizing the marginal entropy across augmented views. Sun \textit{et al.}~\cite{sun2020test} pioneered self-supervised TTA by reformulating each unlabeled test sample as a self-supervised learning task, while TTT-MAE~\cite{gandelsman2022test} employs masked autoencoders~\cite{he2022masked} for per-sample self-supervision, allowing the model to adjust dynamically to each test input.
More recently, several works have focused on improving the robustness of TTA methods.
NOTE~\cite{gong2022note} and RoTTA~\cite{yuan2023robust} investigate TTA performance under non-i.i.d. and temporally correlated test streams, while OSTTA~\cite{dong2025towards} addresses performance degradation caused by long-term adaptation.
Notably, existing TTA research primarily targets feature-level semantic  shifts, whereas our work focuses on the more challenging semantic or label semantic shifts.
Although a few studies have explored scenarios where test data contain unknown categories, they either do not explicitly distinguish among the unknown classes~\cite{li2023robustness,dong2025towards} or require access to unlabeled samples from new categories during training~\cite{rathore2025domain,rongali2024cdad,wang2024hilo}.

% Notably, OTTA mainly focuses on addressing semantic
% shifts in the feature domain while we aim to resolve the challenge of label semantic shift caused by
% unseen compositions recombined from attributes and objects in CZSL. Similarly, the CZSL task does
% not have access to unseen compositions during training.
\section{The Proposed Method}

\begin{figure*}[t]
  \centering
  \includegraphics[width=0.85\textwidth]{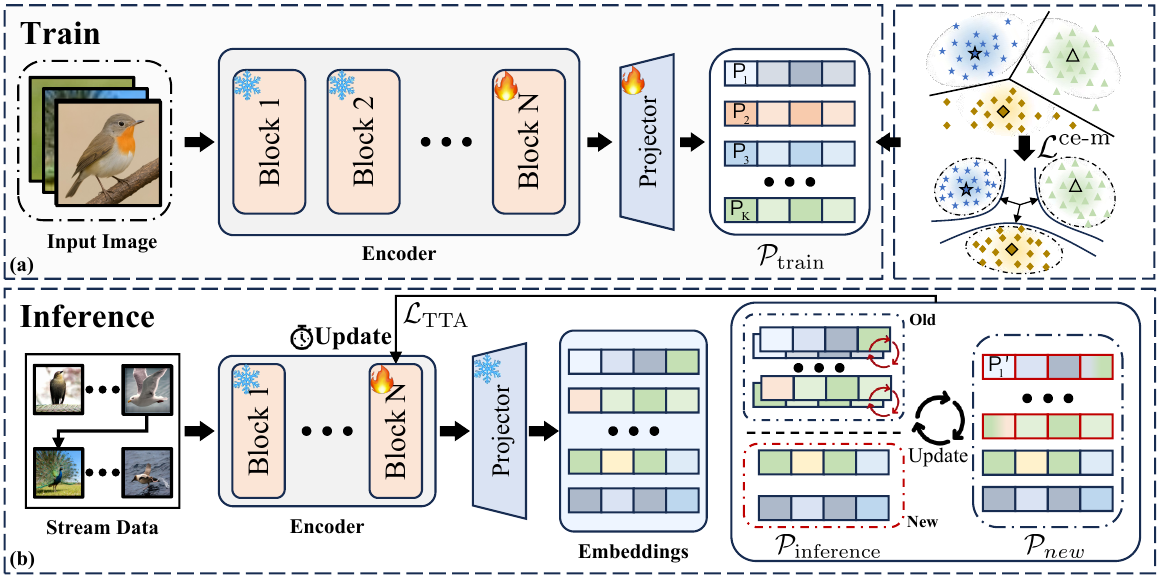}
  \caption{\textbf{Overview of the proposed TALON framework.} (a) During the offline stage, we introduce margin-aware logit calibration to enlarge inter-class margins and enhance intra-class compactness, reserving embedding space for future category discovery. (b) At test-time, we jointly update the encoder and class prototypes, enabling the model to learn through discovery rather than static inference.
  }
  \label{fig:framework}
  \vspace{-5mm}
\end{figure*}

\paragraph{Problem definition.} 
On-the-fly category discovery (OCD) aims to identify both known and novel categories from an incoming data stream in an online manner, relying solely on the knowledge of labeled samples from known classes during offline training. Formally, we are provided with a labeled support set denoted as $\mathcal{D}_S = \{(\mathbf{x}_i, y_i) \}_{i=1}^{M} \subseteq \mathcal{X} \times \mathcal{Y}_S$ for training, and an unlabeled query set denoted as $\mathcal{D}_Q = \{(\mathbf{x}_i, y_i)\}_{i=M+1}^{M+N} \subseteq \mathcal{X} \times \mathcal{Y}_Q$ for testing 
% \yang{make it explicit that $y_i$ in query set is unobserved. }
, where the labels $y_i$ in $\mathcal{D}_Q$ are inaccessible during inference and used only for evaluation. $M$ and $N$ represent the numbers of samples in $\mathcal{D}_S$ and $\mathcal{D}_Q$, respectively. 
$\mathcal{Y}_S$ and $\mathcal{Y}_Q$ represent the label spaces of known and all test classes, with $\mathcal{Y}_S \subseteq \mathcal{Y}_Q$.  We refer to classes in $\mathcal{Y}_S$ as known/old categories and those in $\mathcal{Y}_Q \setminus \mathcal{Y}_S$ as novel/unknown categories. 
 Only the support set $\mathcal{D}_S$ is available for model training, while the query samples arrive sequentially during testing.

 %\yang{maybe mention that we do not have access to the support set after offline training}
 
% \noindent 
In this work, we take a step beyond prior OCD formulations by introducing a \emph{test-time adaptation} (TTA) framework into the task. 
By leveraging the discriminative structure learned from labeled known classes, our model continually self-adjusts during inference to handle semantic-aware shifts (e.g., the emergence of novel categories) in $\mathcal{D}_Q$.

\paragraph{Methods overview.} 
Motivated by the static inference and limited expressiveness of prior hash-based OCD methods, we propose a TTA framework built upon a hash-free architecture.
It constructs a discriminative and robust representation space directly from visual embeddings and enables the model to learn through discovery continually. Fig.~\ref{fig:framework} shows our overall framework. During offline training, we first learn a model initialization with a margin-aware logit calibration module, which enhances feature compactness and class separability for future category discovery (see Fig.~\ref{fig:framework} (a)).
During online inference, instance-level predictions are made by measuring the feature similarity between the test sample and the prototype memory. Afterwards, the model encoder and prototype memory are updated using the visual features and predictions of incoming test samples, enabling continuous adaptation to the evolving data distribution (see Fig.~\ref{fig:framework} (b)). Notably, our framework provides instant feedback for each incoming instance, while model and prototype updates occur periodically, ensuring a balance between real-time responsiveness and stable adaptation.
% \yang{mention that this is done after each example, otherwise people might thinking we first make predictions for a batch, then do the update} 
In the following, we describe these two parts in detail.

% \subsection{Representation learning via prototype network}

\subsection{Representation learning on labeled data}

To learn a robust and semantically meaningful representation from labeled data, we employ both the supervised contrastive loss \cite{khosla2020supervised} and cross-entropy loss \cite{mao2023cross}. Let $\mathbf{x}_i$ and $\mathbf{x}^{\prime}_{i}$ denote two randomly augmented views of the $i$-th instance sample, the supervised contrastive loss is defined as:
\begin{equation}
\mathcal{L}_i^{\text{sup}}
= -\frac{1}{|\mathcal{N}(i)|}
\sum_{q \in \mathcal{N}(i)}
\log \frac{\exp(\tilde{\mathbf{z}}_i \cdot \tilde{\mathbf{z}}_q / \tau)}
{\sum_{n} \mathbb{I}_{[n \neq i]}\exp(\tilde{\mathbf{z}}_i \cdot \tilde{\mathbf{z}}_n / \tau)},
\label{eq:sup_loss}
\end{equation}
where $\tilde{\mathbf{z}}_i = \mathbf{z}_i / \|\mathbf{z}_i\|_2$ is the $\ell_2$-normalized feature embedding of $\mathbf{x}_i$ produced by the encoder $f$, and $\tau$ is a temperature parameter. 
$\mathcal{N}(i)$ denotes the set of indices of samples sharing the same label as $\mathbf{x}_i$ within the batch (including its augmented view $\mathbf{x}^{\prime}_{i}$), and $\mathbb{I}_{[n \neq i]}$ is an indicator function that equals $1$ when $n \neq i$ and $0$ otherwise.

In addition, we adopt a cross-entropy loss to enhance class-level discrimination further. 
Specifically, a linear projection head is applied on top of $\tilde{\mathbf{z}}_i$ to produce a logit vector $\mathbf{p}_i \in \mathbb{R}^{|\mathcal{Y}_S|}$ over the known categories:
\begin{equation}
\mathbf{p}_i = W\,\tilde{\mathbf{z}}_i + \mathbf{b},
\end{equation}
where $W \in \mathbb{R}^{|\mathcal{Y}_S|\times d}$ and $\mathbf{b}$ are the learnable weights and bias of the projector, with $|\mathcal{Y}_S|$ denoting the number of labeled classes and $d$ the feature dimension; each class vector corresponds to class $c$.
The cross-entropy loss is then given by
\begin{equation}
\mathcal{L}_i^{\text{ce}} 
= - \log \frac{\exp(p_{i,y_i})}{\sum_{c=1}^{|\mathcal{Y}_S|} \exp(p_{i,c})},
\label{eq:ce_loss}
\end{equation}
where $p_{i,y_i}$ is the logit corresponding to the ground-truth label $y_i$.
Finally, the two losses are combined with a weighting coefficient $\lambda \in [0,1]$ over a mini-batch $B$:
\begin{equation}
\mathcal{L}_{\text{labeled}} 
= \!\sum_{i \in B}\! \mathcal{L}^{\text{sup}}_i 
+ \lambda \!\sum_{i \in B}\! \mathcal{L}^{\text{ce}}_i .
\label{eq:total_labeled}
\end{equation}

% 尽管以上的方法已经可以实现一个可比较的性能，但是考虑到OCD的目标是让模型去发现与已知类别 在 语义上相近且未知的类别，因此，我们设计了一个 XXX 模块，使得不同类别之间有更大的间隔，相同类别内部更加的聚拢。
\paragraph{Margin-aware Logit Calibration.}
Although the above scheme achieves competitive performance, the OCD objective further requires the model to discover novel categories that are \emph{semantically close} to the known ones. 
To this end, we introduce a margin-aware logit calibration that applies a small angular margin to the cosine similarity between $\ell_2$-normalized features and class weights, thereby enlarging inter-class separation while tightening intra-class compactness. 
Here, the class weights correspond to the column vectors of the projector weight matrix $W$ (i.e., $\mathbf{w}_c$ is the $c$-th column of $W$), each representing the prototype direction of a known category in the embedding space.
 Concretely, let $\tilde{\mathbf{z}}_i$ and $\tilde{\mathbf{w}}_c = \mathbf{w}_c/\|\mathbf{w}_c\|_2$ denote the normalized feature and class weight, and define 
\begin{equation}
\cos\theta_{i,c}=\tilde{\mathbf{w}}_c^\top \tilde{\mathbf{z}}_i .
\end{equation}
We then replace the conventional logits with margin-adjusted angular logits:
\begin{equation}
\ell_{i,c}=
\begin{cases}
s\,\cos\!\left(\theta_{i,y_i}+m\right), & \text{if } c=y_i,\\[2pt]
s\,\cos\theta_{i,c}, & \text{if } c\neq y_i,
\end{cases}
\label{eq:margin_logits}
\end{equation}
where $s>0$ is a scaling factor and $m\ge 0$ is an additive angular margin. 
The resulting cross-entropy is
\begin{equation}
\mathcal{L}_i^{\text{ce-m}}
= -\log\frac{\exp(\ell_{i,y_i})}{\sum_{c=1}^{|\mathcal{Y}_S|}\exp(\ell_{i,c})}.
\label{eq:margin_ce}
\end{equation}
By enlarging the angular gap among classes and pulling same-class samples closer, this calibration strengthens class-level separability of known categories and, in turn, facilitates the discovery of semantically proximate yet previously unseen categories during online inference. Therefore, the final training loss is 
$\mathcal{L}_{\text{labeled}}=\mathcal{L}^{\text{sup}}+ \lambda \mathcal{L}^{\text{ce-m}} $.

% 备选：Learning Test-Time Adaptation toward Online Category Discovery
\subsection{Test-Time adaptive online category discovery}

Our offline training provides a strong initialization, which leaves room for learning novel concepts. However, a frozen model faces a nonstationary test stream where category semantics and data statistics drift over time. 
To remain reliable, the model should adapt during test time through periodic updates rather than staying fixed.
Motivated by this need, we couple \emph{an online decision rule} for recognition and novelty handling with two lightweight adaptation mechanisms—\emph{semantic-aware refinement of the prototype memory} and a \emph{stabilized test-time update of the encoder}—thereby maintaining robustness to semantic shift, i.e., the continual emergence of novel categories.

\subsubsection{Online inference and novelty decision}

After training on labeled data, we construct a class-wise \emph{prototype memory} for known categories. 
Specifically, we extract visual features for all labeled samples and compute the mean feature of each class as its prototype.
The prototype of class $c \in \mathcal{Y}_S$ is given by
\begin{equation}
\boldsymbol{\mu}_c = \mathrm{normalize}\!\left(\frac{1}{|\mathcal{I}_c|}\sum_{i \in \mathcal{I}_c} \tilde{\mathbf{z}}_i\right),
\qquad 
\mathcal{I}_c = \{\,i \mid y_i = c\,\},
\end{equation}
and the collection of these prototypes forms the initial memory 
$\mathcal{P} = \{(c, \boldsymbol{\mu}_c)\}_{c \in \mathcal{Y}_S}$.
During online inference, each incoming test sample $\mathbf{x}$ is embedded into a normalized feature $\tilde{\mathbf{z}}$, which is compared with all prototypes in the current memory $\mathcal{P}$ via cosine similarity:
\begin{equation}
s_c = \tilde{\mathbf{z}}^\top \boldsymbol{\mu}_c, \qquad (c,\boldsymbol{\mu}_c) \in \mathcal{P}.
\end{equation}

Let $c^{\ast} = \arg\max_{c} s_c$ and $s_{\max} = s_{c^\ast}$. 
If $s_{\max} \ge \tau$ (a predefined similarity threshold), the sample is classified as the corresponding category, $\hat{y} = c^\ast$. 
Otherwise, it is regarded as an instance of a novel category. 
A new prototype is then created and initialized with its feature representation:
\begin{equation}
\boldsymbol{\mu}_{c_{\text{new}}} \leftarrow \tilde{\mathbf{z}}, \qquad 
\mathcal{P} \leftarrow \mathcal{P} \cup \{(c_{\text{new}}, \boldsymbol{\mu}_{c_{\text{new}}})\}, \qquad
\hat{y} = c_{\text{new}}.
\label{eq:proto_init}
\end{equation}

This process is applied sequentially to all incoming samples, enabling the model to recognize known categories while dynamically expanding the prototype memory to accommodate newly discovered ones.

\subsubsection{Semantic-aware prototype update}

To ensure that the prototype memory remains representative as new data arrive, we design a \emph{semantic-aware prototype update} mechanism. 
The rationale for continuously evolving class representations is strongly supported by recent advancements in on-the-fly and continual category discovery~\cite{NCD-DLT, GM, wu2023metagcd}, which highlight the critical need for adaptive mechanisms to handle dynamic and potentially long-tailed data streams. 
Guided by these principles, each prototype is dynamically refined during online inference based on the incoming samples assigned to it.
Let $\mathcal{S}_j=\{\tilde{\mathbf{z}}_1,\ldots,\tilde{\mathbf{z}}_{n_j}\}$ denote the features associated with the $j$-th prototype $\boldsymbol{\mu}_j$. 
We first compute their normalized mean feature 
$\bar{\mathbf{z}}_j = \mathrm{normalize}\!\left(\frac{1}{n_j}\sum_{i=1}^{n_j}\tilde{\mathbf{z}}_i\right)$
and an average cosine similarity to measure assignment confidence:
\begin{equation}
\mathrm{conf}_j = \frac{1}{n_j}\sum_{i=1}^{n_j} (\tilde{\mathbf{z}}_i^\top \boldsymbol{\mu}_j), \qquad 
\mathrm{conf}_j \in [0,1].
\end{equation}
The prototype is then updated through an exponential moving average:
\begin{equation}
\boldsymbol{\mu}_j \leftarrow 
\mathrm{normalize}\!\big((1-\alpha_j)\,\boldsymbol{\mu}_j + \alpha_j\,\bar{\mathbf{z}}_j\big),
\label{eq:proto_update}
\end{equation}
where the adaptive step size $\alpha_j$ is defined as
\begin{equation}
\alpha_j = \eta \cdot \mathrm{conf}_j \cdot \frac{n_j}{n_j + \kappa},
\label{eq:alpha_adapt} % <--- Reviewer提到的 Eq. 13
\end{equation}
with $\eta$ as the base update rate and $\kappa$ a smoothing constant.  
Crucially, this formulation specifically addresses the numerical stability of the online inference process. The fractional term $\frac{n_j}{n_j + \kappa}$ controls the \emph{effective sample size}, naturally encoding two conservative update principles:  
(1) Prototypes are updated more aggressively only when they have high confidence and are supported by a substantial number of samples. 
(2) The update magnitude is heavily down-weighted when data are scarce ($n_j \ll \kappa$) or confidence is low. 
This mechanism is vital for mitigating instability arising from extreme data orderings; even if an early outlier temporarily triggers the creation of a pseudo-class via Eq.~\eqref{eq:proto_init}, its lack of subsequent supporting samples ensures it yields negligible updates. This prevents noisy drift and stops outlier prototypes from persisting as dominant categories. 
After each update, all prototypes are re-normalized to remain on the unit hypersphere:
\begin{equation}
\boldsymbol{\mu}_j \leftarrow \boldsymbol{\mu}_j / \|\boldsymbol{\mu}_j\|_2.
\end{equation}

In this way, the prototype memory evolves in a semantic-aware and confidence-controlled manner, maintaining clear decision boundaries and allowing effective tracking of newly emerging categories during online evaluation.

\begin{algorithm}[t]
\caption{Online inference via adaptive update}
\label{alg:online_tta}
\KwIn{Trained encoder $f(\cdot; \theta)$; prototype memory 
$\mathcal{P}=\{(c,\boldsymbol{\mu}_c)\}_{c\in\mathcal{Y}_S}$; 
threshold $\tau$; update rates $\eta,\gamma,\beta_1,\beta_2$; smoothing constant $\kappa$; 
test stream $\mathcal{D}_Q$.}
\KwOut{Updated encoder $\theta$ and refined prototypes $\mathcal{P}$.}

\For{each batch $\mathcal{B}=\{\mathbf{x}_i\}$ from $\mathcal{D}_Q$}{
    Initialize temporary assignment sets $\{\mathcal{S}_j\}$;\\
    \For{each $\mathbf{x}_i \in \mathcal{B}$}{
        $\tilde{\mathbf{z}}_i \!\leftarrow\! f(\mathbf{x}_i; \theta)/\|f(\mathbf{x}_i; \theta)\|_2$;\\
        Compute $s_c=\tilde{\mathbf{z}}_i^\top\boldsymbol{\mu}_c$, $(c,\boldsymbol{\mu}_c)\!\in\!\mathcal{P}$; 
        $c^\ast=\arg\max_c s_c$;\\
        \eIf{$s_{c^\ast}\!\ge\!\tau$}{
            Assign $\hat{y}_i=c^\ast$, add $\tilde{\mathbf{z}}_i$ to $\mathcal{S}_{c^\ast}$;
        }{
            Create new prototype $\boldsymbol{\mu}_{\text{new}}\!\leftarrow\!\tilde{\mathbf{z}}_i$ and 
            $\mathcal{P}\!\leftarrow\!\mathcal{P}\!\cup\!\{(c_{\text{new}},\boldsymbol{\mu}_{\text{new}})\}$;
        }
    }

    \For{each prototype $\boldsymbol{\mu}_j$ with $\mathcal{S}_j$}{
        $\mathrm{conf}_j\!=\!\frac{1}{|\mathcal{S}_j|}\!\sum_{\tilde{\mathbf{z}}_i\in\mathcal{S}_j}\!
        (\tilde{\mathbf{z}}_i^\top\boldsymbol{\mu}_j)$; 
        $\alpha_j\!=\!\eta\,\mathrm{conf}_j\,\frac{|\mathcal{S}_j|}{|\mathcal{S}_j|+\kappa}$;\\
        $\boldsymbol{\mu}_j\!\leftarrow\!\mathrm{normalize}\big((1-\alpha_j)\boldsymbol{\mu}_j+\alpha_j\bar{\mathbf{z}}_j\big)$;
    }

    Compute $\mathcal{L}_{\text{TTA}}\!=\mathcal{L}_{\text{ent}}
+ \beta_1 \mathcal{L}_{\text{align}}
+ \beta_2 \mathcal{L}_{\text{sep}}$;\\
    Update encoder: $\theta\!\leftarrow\!\theta-\gamma\nabla_\theta\mathcal{L}_{\text{TTA}}$;
}
\end{algorithm}

% \vspace{-15mm}

\begin{table*}[ht]
\centering
\small
\setlength{\tabcolsep}{5pt}
\renewcommand{\arraystretch}{1.10} 
\caption{\textbf{Comparison with other SOTA methods.} “DiffGRE+S” and “DiffGRE+P” represent the integration of DiffGRE with SMILE and PHE, respectively.
The best results are highlighted in \textbf{bold}, while the second-best results are \underline{underlined}. \textcolor{blue}{Blue} numbers indicate results not reported in the original papers but reproduced using the official code.}
\resizebox{\textwidth}{!}{
\begin{tabular}{M{1mm} M{24mm}
|ccc|ccc|ccc|ccc|ccc|ccc|ccc}
\toprule
& \textbf{Method}
& \multicolumn{3}{c|}{\textbf{CIFAR10 (\%)}} 
& \multicolumn{3}{c|}{\textbf{CIFAR100 (\%)}} 
& \multicolumn{3}{c|}{\textbf{ImageNet-100 (\%)}}
& \multicolumn{3}{c|}{\textbf{CUB-200-2011 (\%)}} 
& \multicolumn{3}{c|}{\textbf{Stanford Cars (\%)}} 
& \multicolumn{3}{c|}{\textbf{Oxford Pets (\%)}} 
& \multicolumn{3}{c}{\textbf{Food101 (\%)}} \\
&
& All & Old & New
& All & Old & New
& All & Old & New
& All & Old & New
& All & Old & New
& All & Old & New
& All & Old & New \\
\midrule
% =================== GREEDY BLOCK ===================
\multirow{7}{*}{\rotatebox[origin=c]{90}{\textit{Greedy--Hungarian}}}
& SLC \cite{hartigan1975clustering}
& 65.9 & 96.5 & 50.9
& 46.9 & 62.1 & 16.6
& 34.2 & 86.6 & 7.1
& 30.2 & 46.5 & 22.1
& 14.4 & 23.9 & 9.8
& -- & -- & --
& -- & -- & -- \\
& MLDG \cite{MLDGli2018learning}
& 71.6 & \uline{97.5} & 58.6
& 58.4 & 69.0 & 37.3
& 33.6 & 74.4 & 13.1
& 34.2 & 57.9 & 22.4
& 28.0 & 49.1 & 17.7
& -- & -- & --
& -- & -- & -- \\
& RankStat \cite{RankStat}
& 56.5 & 81.1 & 44.2
& 36.9 & 45.7 & 19.3
& 33.1 & 74.2 & 12.4
& 22.7 & 30.0 & 19.1
& 16.2 & 23.3 & 12.8
& -- & -- & --
& -- & -- & --- \\
& WTA \cite{WTAjia2021joint}
& 65.4 & 88.0 & 54.1
& 44.1 & 55.5 & 21.2
& 33.1 & 75.8 & 11.7

& 23.8 & 30.5 & 20.4
& 18.3 & 26.2 & 14.5
& -- & -- & --
& -- & -- & -- \\
& SMILE \cite{SMILE}
& 78.2 & \textbf{99.3} & 67.6
& 61.3 & 70.7 & 42.5
& 39.9 & 87.1 & 16.2

& 41.1 & 67.6 & 27.8
& 33.4 & 58.4 & 21.3
& \textcolor{blue}{54.1} &\textcolor{blue}{66.1}  & \textcolor{blue}{47.8}
& \textcolor{blue}{34.4}   &\textcolor{blue}{64.0}   & \textcolor{blue}{19.4} \\
& \textbf{Ours-DINO}
& \cellcolor{ltgray}{\textbf{86.2}} & \cellcolor{ltgray}{95.4} & \cellcolor{ltgray}{\textbf{79.3}}
& \cellcolor{ltgray}{\textbf{72.5}} & \cellcolor{ltgray}{\textbf{85.2}} & \cellcolor{ltgray}{\textbf{47.0}}
& \cellcolor{ltgray}{\textbf{84.1}} & \cellcolor{ltgray}{\uline{94.3}} & \cellcolor{ltgray}{\textbf{63.4}}
& \cellcolor{ltgray}{\uline{52.6}} & \cellcolor{ltgray}{\uline{83.3}} & \cellcolor{ltgray}{\uline{37.2}}
& \cellcolor{ltgray}{\uline{42.9}} & \cellcolor{ltgray}\uline{78.1} & \cellcolor{ltgray}\uline{25.9}
& \cellcolor{ltgray}\textbf{81.0}   & \cellcolor{ltgray}\uline{92.4}   & \cellcolor{ltgray}\textbf{75.1}
& \cellcolor{ltgray}\uline{44.5}   & \cellcolor{ltgray}\uline{80.4}   & \cellcolor{ltgray}\uline{26.2} \\
& \textbf{Ours-CLIP}
& \cellcolor{ltgray}\uline{84.7} & \cellcolor{ltgray}96.0 & \cellcolor{ltgray}\uline{76.3}
& \cellcolor{ltgray}\uline{69.9} & \cellcolor{ltgray}\uline{82.8} & \cellcolor{ltgray}\uline{44.2}
& \cellcolor{ltgray}\uline{83.6} & \cellcolor{ltgray}\textbf{96.1} & \cellcolor{ltgray}\uline{58.2}
& \cellcolor{ltgray}\textbf{58.9} & \cellcolor{ltgray}\textbf{87.3} & \cellcolor{ltgray}\textbf{44.7}
& \cellcolor{ltgray}\textbf{60.4} & \cellcolor{ltgray}\textbf{90.6} & \cellcolor{ltgray}\textbf{45.8}
& \cellcolor{ltgray}\uline{74.9} & \cellcolor{ltgray}\textbf{94.6} & \cellcolor{ltgray}\uline{64.6}
& \cellcolor{ltgray}\textbf{61.2} & \cellcolor{ltgray}\textbf{88.3} & \cellcolor{ltgray}\textbf{47.3} \\
\midrule
% =================== STRICT BLOCK ===================
\multirow{10}{*}{\rotatebox[origin=c]{90}{\textit{Strict--Hungarian}}}
& SLC \cite{hartigan1975clustering}
& 41.5 & \textbf{58.3} & 33.3
& 44.4 & 59.0 & 15.1
& 32.9 &86.6 & 5.2
& 28.6 & 44.0 & 20.9
& 14.0 & 23.0 & 9.7
& 35.5 & 41.3 &33.1
& 20.9 &48.6  &6.8 \\
& MLDG \cite{MLDGli2018learning}
& 44.1 & 38.5 & 47.0
& 50.6 & 61.0 & 29.8
& 30.6 & 72.3 & 9.7
& 29.5 & 48.4 & 20.1
& 24.0 & 41.6 & 15.4
& --   & --   & --
& --   & --   & -- \\
& RankStat \cite{RankStat}
& 42.1 & \uline{49.3} & 38.6
& 35.0 & 44.0 & 17.0
& 31.1 & 73.3 & 9.8
& 21.2 & 26.9 & 18.4
& 14.8 & 19.9 & 12.3
& 33.2 & 42.3 & 28.4
& 22.3 & 50.7 &7.8 \\
& WTA \cite{WTAjia2021joint}
& 43.1 & 34.5 & 47.4
& 40.8 & 52.9 & 16.7
& 30.8 & 72.9 & 9.7
& 21.9 & 26.9 & 19.4
& 17.1 & 24.4 & 13.6
& 35.2 & 46.3 & 29.3
& 18.2 & 40.5 & 6.1 \\
& SMILE \cite{SMILE}
& 49.9 & 39.9 & 54.9
& 51.6 & 61.6 & 31.7
& 33.8 & 74.2 & 13.5
& 32.2 & 50.9 & 22.9
& 26.2 & 46.6 & 16.3
& 41.2 & 42.1 & 40.7
& 24.0 & 54.6 & 8.4 \\
& PHE \cite{PHE}
& \textcolor{blue}{53.1} &\textcolor{blue}{19.3} &\textcolor{blue}{\uline{70.0}}
& \textcolor{blue}{56.0} & \textcolor{blue}{70.1} & \textcolor{blue}{27.8}
& \textcolor{blue}{39.2} &\textcolor{blue}{49.3} &\textcolor{blue}{34.1}
& 36.4 & 55.8 & 27.0
& 31.3 & 61.9 & 16.8
& 48.3 & 53.8 & 45.4
& 29.1 & \uline{64.7} & 11.1 \\ %our run results
& DiffGRE+S \cite{DiffGRE}
& -- & -- & --
& -- & -- & --
& -- & -- & --
& 35.4 & 58.2 & 23.8
& 30.5 & 59.3  &16.5 
& 42.4 & 42.1  &42.5
& --   & --   & -- \\
& DiffGRE+P \cite{DiffGRE}
& -- & -- & --
& -- & -- & --
& -- & -- & --
& 37.9 & 57.0 & 28.3
& 32.1 & 63.3 & 16.9
& 48.6 & 52.6 & 46.6
& --   & --   & -- \\

& \textbf{Ours-DINO}
& \cellcolor{ltgray}{\textbf{65.0}} & \cellcolor{ltgray}{46.1} & \cellcolor{ltgray}{\textbf{79.3}}
& \cellcolor{ltgray}{\textbf{64.7}} & \cellcolor{ltgray}{\textbf{77.4}} & \cellcolor{ltgray}{\textbf{39.3}}
& \cellcolor{ltgray}{\textbf{82.6}} & \cellcolor{ltgray}{\uline{92.0}} & \cellcolor{ltgray}{\textbf{63.4}}
& \cellcolor{ltgray}{\uline{43.6}} & \cellcolor{ltgray}{\textbf{63.5}} & \cellcolor{ltgray}{\uline{33.6}}
& \cellcolor{ltgray}{\uline{37.0}} & \cellcolor{ltgray}{\uline{68.1}} & \cellcolor{ltgray}{\uline{22.0}}
& \cellcolor{ltgray}{\textbf{69.2}}   & \cellcolor{ltgray}{\uline{58.5}}   & \cellcolor{ltgray}{\textbf{74.8}}
& \cellcolor{ltgray}{\uline{30.3}}   & \cellcolor{ltgray}{60.5}   & \cellcolor{ltgray}{\uline{15.0}} \\
& \textbf{Ours-CLIP}
& \cellcolor{ltgray}\uline{56.9} & \cellcolor{ltgray}31.1 & \cellcolor{ltgray}\uline{76.3}
& \cellcolor{ltgray}\uline{61.6} & \cellcolor{ltgray}\uline{75.0} & \cellcolor{ltgray}\uline{34.9}
& \cellcolor{ltgray}\uline{80.9} & \cellcolor{ltgray}\textbf{93.6} & \cellcolor{ltgray}\uline{54.9}
& \cellcolor{ltgray}\textbf{45.5} & \cellcolor{ltgray}\uline{60.7} & \cellcolor{ltgray}\textbf{37.8}
& \cellcolor{ltgray}\textbf{53.5} & \cellcolor{ltgray}\textbf{74.2} & \cellcolor{ltgray}\textbf{43.6}
& \cellcolor{ltgray}\uline{64.0} & \cellcolor{ltgray}\textbf{65.4} & \cellcolor{ltgray}\uline{63.3}
& \cellcolor{ltgray}\textbf{50.3} & \cellcolor{ltgray}\textbf{66.2} & \cellcolor{ltgray}\textbf{42.2} \\
\bottomrule
\end{tabular}
}
\label{acc}
\vspace{-1.5em}
\end{table*}
%\vspace*{-2cm}
\subsubsection{Stable test-time parameter adaptation}
While prototype refinement maintains semantic consistency, the encoder itself should also adapt to semantical changes in the test stream without overfitting. 
To this end, the model periodically collects a small batch of recent unlabeled samples and performs a few gradient updates on the encoder parameters, guided by an entropy-minimization objective with semantic regularization.

During each adaptation period, visual features $\{\tilde{\mathbf{z}}_i\}_{i=1}^{N}$ are extracted from incoming test samples, and their similarities to prototypes $(c,\boldsymbol{\mu}_c)\in\mathcal{P}$ are computed as $s_c = \tilde{\mathbf{z}}_i^\top \boldsymbol{\mu}_c$. The resulting similarities are normalized by a softmax function $p_i(c) = \frac{\exp(s_c / T)}{\sum{c'} \exp(s{c'} / T)}$, where $c'$ indexes all prototypes in $\mathcal{P}$ and $T$ is a temperature controlling confidence sharpness.
To enable stable and semantically consistent adaptation, we define an entropy-based loss augmented with prototype-level regularization. 
First, we minimize the average predictive entropy to encourage confident predictions:
\begin{equation}
\mathcal{L}_{\text{ent}} = -\frac{1}{N}\sum_{i=1}^{N}\sum_{c=1}^{|\mathcal{P}|} p_i(c)\,\log p_i(c),
\end{equation}
which refines decision boundaries by driving features toward low-uncertainty regions.  
Meanwhile, to preserve semantic consistency between features and their corresponding prototypes, we align each class mean feature with its stored prototype. 
Let $\bar{\mathbf{z}}_l = \mathrm{normalize}\!\left(\frac{1}{|\mathcal{S}_l|}\sum_{i\in\mathcal{S}_l}\tilde{\mathbf{z}}_i\right)$ denote the $\ell_2$-normalized mean embedding of class $l$. 
The alignment loss is defined as:
\begin{equation}
\mathcal{L}_{\text{align}} = -\,\frac{1}{|\mathcal{P}|}\sum_{l} 
\big(\bar{\mathbf{z}}_l^\top \boldsymbol{\mu}_l\big).
\end{equation}

To maintain inter-class discriminability, we further penalize excessive similarity between different class means:
\begin{equation}
\mathcal{L}_{\text{sep}} = \sum_{l \ne m}\big(1 - \bar{\mathbf{z}}_l^\top \bar{\mathbf{z}}_m\big),
\end{equation}
which prevents cluster collapse and encourages clear class boundaries.
The overall adaptation objective is given by
\begin{equation}
\mathcal{L}_{\text{TTA}} =
\mathcal{L}_{\text{ent}}
+ \beta_1 \mathcal{L}_{\text{align}}
+ \beta_2 \mathcal{L}_{\text{sep}},
\end{equation}
where $\beta_1$ and $\beta_2$ control the strengths of alignment and separation. 
Gradients are backpropagated only through the encoder parameters, while the prototype memory remains fixed in each adaptation. With these lightweight periodic updates, the encoder stays aligned with evolving prototypes, yielding lower-entropy predictions, better semantic consistency, and clearer class boundaries for robust recognition under continuous semantical shifts.
The pseudo-code for online inference with adaptive updates is provided in Alg.~\ref{alg:online_tta}

% \yang{the parameter adaptation requires a batch of examples, right? Would this violate the OCD assumption? Shall we clarify the setting a bit more?}
\section{Experiment}

\subsection{Dataset and setup}
\paragraph{Datasets.} We conduct experiments on seven benchmark datasets, including three coarse-grained classification datasets: CIFAR10 \cite{cifar}, CIFAR100 \cite{cifar}, ImageNet-100 \cite{2015imagenet}, and four fine-grained datasets: CUB-200-2011 \cite{CUB}, Stanford Cars \cite{scars}, Oxford-IIIT Pet \cite{pets}, and Food-101 \cite{food}. Note that ImageNet-100 refers to a subset of ImageNet with 100 categories randomly sampled. Following the setup in OCD \cite{SMILE} and PHE \cite{PHE}, the categories of each dataset are split into subsets of seen and unseen categories. Specifically, 50\% samples belonging to the seen categories are used to form the labeled set $\mathcal{D}_S$, and 
the rest forms the unlabeled set $\mathcal{D}_Q$ for on-the-fly testing. 
Detailed information about the datasets used is provided in the Appendix.

\vspace*{-0.3cm}
\paragraph{Evaluation Metrics.} 
Following \cite{SMILE}, we adopt two protocols for evaluation termed \textit{Greedy-Hungarian} and \textit{Strict-Hungarian} for comprehensive comparisons, where their difference is clearly illustrated in \cite{vaze2022generalized}.
During testing, samples sharing the same category descriptor form a cluster, and only the top-$|\mathcal{Y}_Q|$ clusters by size are retained, with the rest treated as misclassified.
For \textit{Greedy-Hungarian}, accuracy is computed separately on the “Old” and “New” subsets, providing independent evaluations of known and novel classes.
In contrast, \textit{Strict-Hungarian} calculates accuracy over the entire query set, avoiding repeated cluster assignments between subsets.
The overall accuracy is obtained via the Hungarian matching:
\vspace{-1mm}
 \begin{equation}
ACC=\max\limits_{p\in \mathcal{P}(\mathcal{Y}_Q)}\frac{1}{ \left| \mathcal{D}_Q \right| }\sum^{\left| \mathcal{D}_Q \right|}_{i=1}\mathds{1}[y_i = p(\hat{y}_i)],
\end{equation}
% \vspace{-1mm}
where $y_i$ is the ground truth label, $\hat{y}_i$ is the predicted label decided by cluster indices, and $\mathcal{P}(\mathcal{Y}_Q) $ is the set of all permutations of ground truth lables.

\paragraph{Implementation Details.} 

\vspace*{-0.4cm}
For fair comparison, we follow SMILE~\cite{SMILE} and adopt DINO~\cite{caron2021emerging} pre-trained ViT-B-16~\cite{dosovitskiy2020image} as the backbone.
We also report results using CLIP~\cite{radford2021learning} pre-trained ViT-B-16, which generally achieves stronger performance across datasets (see Section~\ref{sec:sota}).
During training, only the final transformer block is fine-tuned, and the projector is a single linear layer whose output dimension equals $|\mathcal{Y}_S|$.
We do not perform extensive hyperparameter tuning and all datasets share almost identical configurations.
The angular margin $m$ in the MLC module is fixed to 0.2, and the adaptation batch size is set to 64. The loss weights in $\mathcal{L}_{\text{labeled}}$ and $\mathcal{L}_{\text{TTA}}$ are set as $\lambda=\beta_1=\beta_2=1$.
With the DINO backbone, both the encoder and projector use a learning rate of $1\times10^{-3}$ and a similarity threshold of $\tau=0.7$.
For CLIP, learning rates are set to $1\times10^{-4}$ for the encoder and $1\times10^{-3}$ for the projector, with $\tau=0.75$. 
At test-time, the encoder is updated using a learning rate of $1\times10^{-4}$ for all datasets.
For prototype refinement, the update rate $\eta$ and smoothing constant $\kappa$ are set to 0.06 and 32 for known classes, and to 0.3 and 8 for newly discovered classes. 
 
\begin{table}[t]
  \setlength{\tabcolsep}{4pt}
  \footnotesize
  \renewcommand{\arraystretch}{0.8}
  \centering
  \caption{\textbf{Ablation study of various components of our TALON on the CUB and Scars datasets.} Here,  MLC, TTA-P, and TTA-M denote the margin-aware logit calibration, prototype update, and model adaptation modules, respectively.}
  \label{tab:ab_com}
  \begin{tabular}{l|ccc|ccc}
    \toprule
    \multirow{2}{*}[-0.8ex]{Method} & \multicolumn{3}{c|}{CUB-200-2011 (\%)} & \multicolumn{3}{c}{Stanford Cars (\%)} \\ %& \multicolumn{3}{c}{Pets (\%)} \\
    \cmidrule(lr){2-4} \cmidrule(lr){5-7} %\cmidrule(lr){8-10}
    & All & Old & New & All & Old & New \\%& All & Old & New\\
    \midrule
    Baseline              
    & 44.5 & 57.6 & 37.9 &47.8 &66.7 & 38.6 \\ %&57.0 &68.0 &51.3 \\
    Baseline+MLC           
    & 45.7 & 58.6 & 39.2 &49.0 &65.4 &41.4 \\ %&61.7 &70.4 & 57.1\\
    Baseline+TTA-P 
    &45.4  &56.4  &39.9  &48.7  &63.0  &41.8 \\ %&54.6 &56.6 &53.6  \\
    Baseline+MLC+TTA-P  
    &46.7  &57.7  &41.2 &52.7  &69.7  &44.5 \\ %&57.4 &60.3 &55.8 \\ 
    Baseline+TTA-M                      
    &44.5  &56.4  &38.5  &47.7  &66.3  &38.8 \\ %& 57.9 &68.0 &52.6  \\
    Baseline+MLC+TTA-M  
    &46.7  &57.8  &41.1  &52.1  &71.1  &42.9 \\ %&57.4 &60.0 &56.0  \\
    Baseline+TTA-P-M  
    &44.4  &56.4  &38.4  &48.6  &67.5  &39.4 \\ %&61.2 &73.3 &54.8 \\
    \rowcolor{gray!10}
    \textbf{TALON(ours)}                  
    & \text45.5 &60.7  &37.8 &53.5 &74.2 &43.6 \\ %&62.2 &70.6 &57.8 \\
    \bottomrule
  \end{tabular}
  \vspace{-5mm}
  \label{tab:ablation}
\end{table}

\begin{table}[t]
% \vspace{-3mm}
\small
\centering
\setlength{\tabcolsep}{2pt}
\renewcommand{\arraystretch}{1.0}
\caption{\textbf{Comparison of estimated category numbers and accuracies on the CUB-200-2011 and Stanford Cars datasets.} }
\label{tab:pred_acc_count}
\begin{tabular}{l|c|ccc|c|ccc}
\toprule
\multirow{2}{*}[-0.6ex]{Method}
& \multicolumn{4}{c|}{CUB ($C{=}200$)} 
& \multicolumn{4}{c}{SCars ($C{=}196$)} \\
\cmidrule(lr){2-5} \cmidrule(lr){6-9}
& \#Cls & All & Old & New & \#Cls & All & Old & New \\
\midrule
SMILE-64bit~\cite{SMILE}    
& 2910 & 22.6 & 45.3 & 11.2 & 4788 & 16.5 & 38.2 & 6.1 \\
SMILE-32bit~\cite{SMILE}    
& 2146 & 27.3 & 52.0 & 15.0 & 2953 & 21.9 & 46.8 & 9.9 \\
SMILE-16bit~\cite{SMILE}    
& 924 & 31.9 & 52.7 & 21.5 & 896 & 27.5 & 52.5 & 15.4 \\
PHE-64bit~\cite{PHE}        
&493  &38.1 &60.1 &27.2 & 917 &32.1 & 66.9 & 15.3 \\
PHE-32bit~\cite{PHE}        
& 474 & 38.5 & 59.9 & 27.8 & 762 & 31.5 & 64.0 & 15.8 \\
PHE-16bit~\cite{PHE}        
& 318 & 37.6 & 57.4 & 27.6 & 709 & 31.8 & 65.4 & 15.6 \\
\rowcolor{gray!10}
\textbf{TALON(ours)} & 
153 
& 45.5 & 60.7 & 37.8 & 299 & 53.5 & 74.2 & 43.6 \\
\bottomrule
\end{tabular}
\label{tab:category_number}

\end{table}

\subsection{Comparison with the state-of-the-art}
\label{sec:sota}
We conduct comparative experiments on the aforementioned seven datasets, with results summarized in Tab.~\ref{acc}. Our method consistently outperforms previous approaches across all datasets and evaluation protocols. We report results using both DINO and CLIP as backbones (i.e., Ours-DINO vs. Ours-CLIP). The two variants perform comparably on coarse-grained datasets, while Ours-CLIP exhibits a clear advantage on fine-grained ones. This improvement can be attributed to CLIP’s stronger visual representations learned from large-scale image–text pretraining, which better capture subtle semantic differences. In contrast, DINO, trained purely on ImageNet, produces more generic features that are less specialized for fine-grained recognition.

In particular, Ours-CLIP significantly outperforms Ours-DINO on fine-grained datasets such as SCars and Food101, achieving 60.4\% and 61.2\%, compared to 42.9\% and 44.5\% in All classes accuracy. Furthermore, on CIFAR10 and ImageNet-100, our DINO-based model attains 86.2\% and 79.3\% on New classes, surpassing previous methods by large margins. The advantage remains consistent on large-scale or challenging datasets such as ImageNet-100 and Food101, where our DINO-based variant still achieves competitive results of 94.3\% and 80.4\% on Old Classes.

\subsection{Ablation study}

\paragraph{The effectiveness of each module.}
Tab.~\ref{tab:ablation} reports the results of the ablation study, illustrating the individual and combined effects of different components in our TALON framework.
Starting from the plain baseline ($\mathcal{L}_{\text{sup}}+\mathcal{L}_{\text{ce}}$ with static inference), the introduction of MLC already brings consistent improvements on both datasets, confirming that enhancing inter-class separability and intra-class compactness during offline training provides a stronger feature foundation for subsequent online category discovery.
Incorporating TTA-P further improves performance by dynamically refining class prototypes as new data arrive, ensuring that the memory remains semantically representative and robust to incremental shifts.
Meanwhile, TTA-M contributes additional gains by directly adapting the encoder parameters to the evolving data distribution, enabling the model to maintain stable and discriminative feature representations at test time.
When both modules are combined (TTA-P-M), the system benefits from the complementary nature of prototype-level and model-level adaptation, leading to further improvement.
Finally, the complete TALON achieves the highest overall performance across both CUB and SCars datasets, demonstrating the effectiveness and strong synergy among all components.

\vspace*{-0.3cm}
\paragraph{Evaluation of estimated category number.}
Tab.~\ref{tab:category_number} reports the estimated number of discovered categories (\#Cls) and accuracy on CUB ($C{=}200$) and SCars ($C{=}196$). 
Hashing-based methods (SMILE and PHE) show strong dependence on the hash code length $L$, as the prediction space grows exponentially with $2^L$. 
Larger $L$ values cause severe overestimation of category numbers, while shorter codes reduce discriminability, leading to poor accuracy, especially for new classes. 
In contrast, our hash-free framework directly models visual features and jointly optimizes them through the proposed modules, effectively mitigating category explosion while achieving accurate and stable category discovery.

% \paragraph{Comparison with existing TTA methods.}

\begin{figure}[t]
    \centering
    \includegraphics[width=1\linewidth]{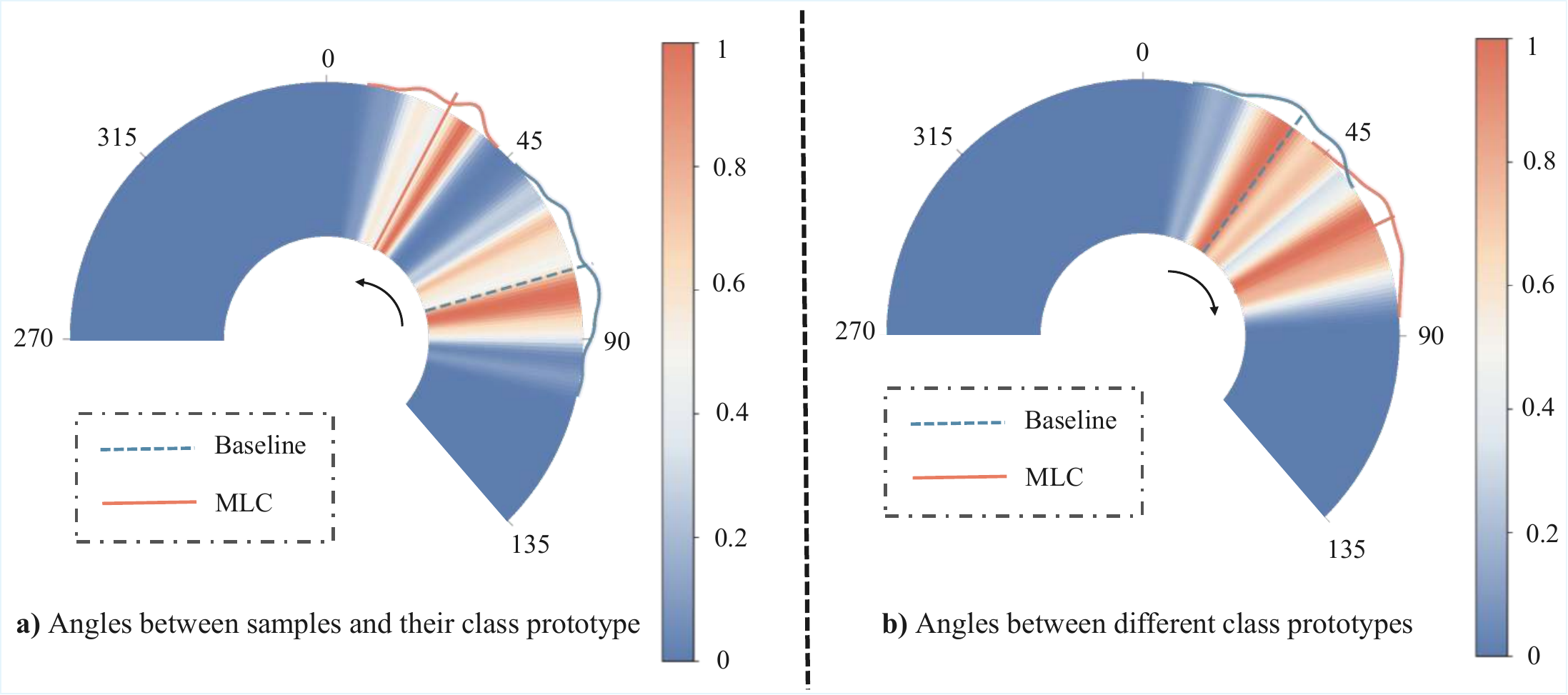}
    \caption{\textbf{Angular analysis of the margin-aware logit calibra-tion on the Pets dataset. } Left: angles between samples and their class prototypes; Right: angles between different class prototypes.
}
    \label{fig:mlc_angle}
    \vspace{-3mm}
\end{figure}

\begin{table}[t]
  \setlength{\tabcolsep}{3.8pt}
  \footnotesize
  \renewcommand{\arraystretch}{0.8}
  \centering
  \caption{\textbf{Comparison with existing test-time adaptation methods on the Stanford Cars and Oxford Pets datasets.}}
  \label{tab:ab_tta}
  \begin{tabular}{l|ccc|ccc}
    \toprule
    \multirow{2}{*}[-0.8ex]{Method} & \multicolumn{3}{c|}{Stanford Cars (\%)} & \multicolumn{3}{c}{Oxford Pets (\%)} \\% & \multicolumn{3}{c}{Pets (\%)} \\
    \cmidrule(lr){2-4} \cmidrule(lr){5-7} % \cmidrule(lr){8-10}
    & All & Old & New & All & Old & New \\% &All & Old & New\\
    \midrule
    baseline+MLC           
    %& 45.7 & 58.6 & 39.2 
    &49.0 &65.4 &41.4 &61.7 &70.4 & 57.1\\
    baseline+MLC+TENT \cite{wang2020tent} 
    %&47.8  &60.3  &41.6  
    &48.1  &66.4  &39.2 &58.4 &66.1 &54.3  \\
    baseline+MLC+OSTTA \cite{dong2025towards}
    %&47.0  &57.6  &41.7  
    &47.2  &62.3  &39.9 &60.3 &70.4 &54.9  \\
    baseline+MLC+TTA-P-TENT 
    %&47.1  &57.5  &41.9  
    &49.4  &67.4  &40.7 &60.0 &65.0 &57.4  \\
    baseline+MLC+TTA-P-OSTTA
    %&46.6  &59.8  &40.0  
    &47.6  &65.2  &39.2 &61.9 &\textbf{70.6} &57.3  \\
    \rowcolor{gray!10}
    \textbf{TALON(ours)}                  
    %& 47.9 &58.5  &42.5 
    &\textbf{53.5} &\textbf{74.2} &\textbf{43.6} &\textbf{64.0} & 65.4 &\textbf{63.3} \\
    \bottomrule
  \end{tabular}
  \vspace{-5mm}
  \label{tab:tta_comparison}
\end{table}

\paragraph{Limitations of existing TTA methods.}
Tab.~\ref{tab:tta_comparison} compares our TALON framework with variants that integrate existing test-time adaptation (TTA) methods, including TENT~\cite{wang2020tent} and OSTTA~\cite{dong2025towards}. 
When applied within our baseline with MLC, these methods bring only marginal improvements, and in some cases even degrade performance. 
This is because conventional TTA approaches are designed for \emph{domain shift}, where the label space is fixed and only the input semantic changes. 
In contrast, our OCD task involves \emph{semantic shift}, which requires not only adapting to new semantic but also discovering unseen categories. TENT and OSTTA struggle to maintain stable recognition under such settings, whereas our TALON explicitly models evolving category semantics through joint adaptation of the prototype memory and the encoder parameters, achieving the best results on both SCars and Pets and demonstrating strong suitability for open-world category discovery.

\begin{figure}[t]
    \centering
    \includegraphics[width=1\linewidth]{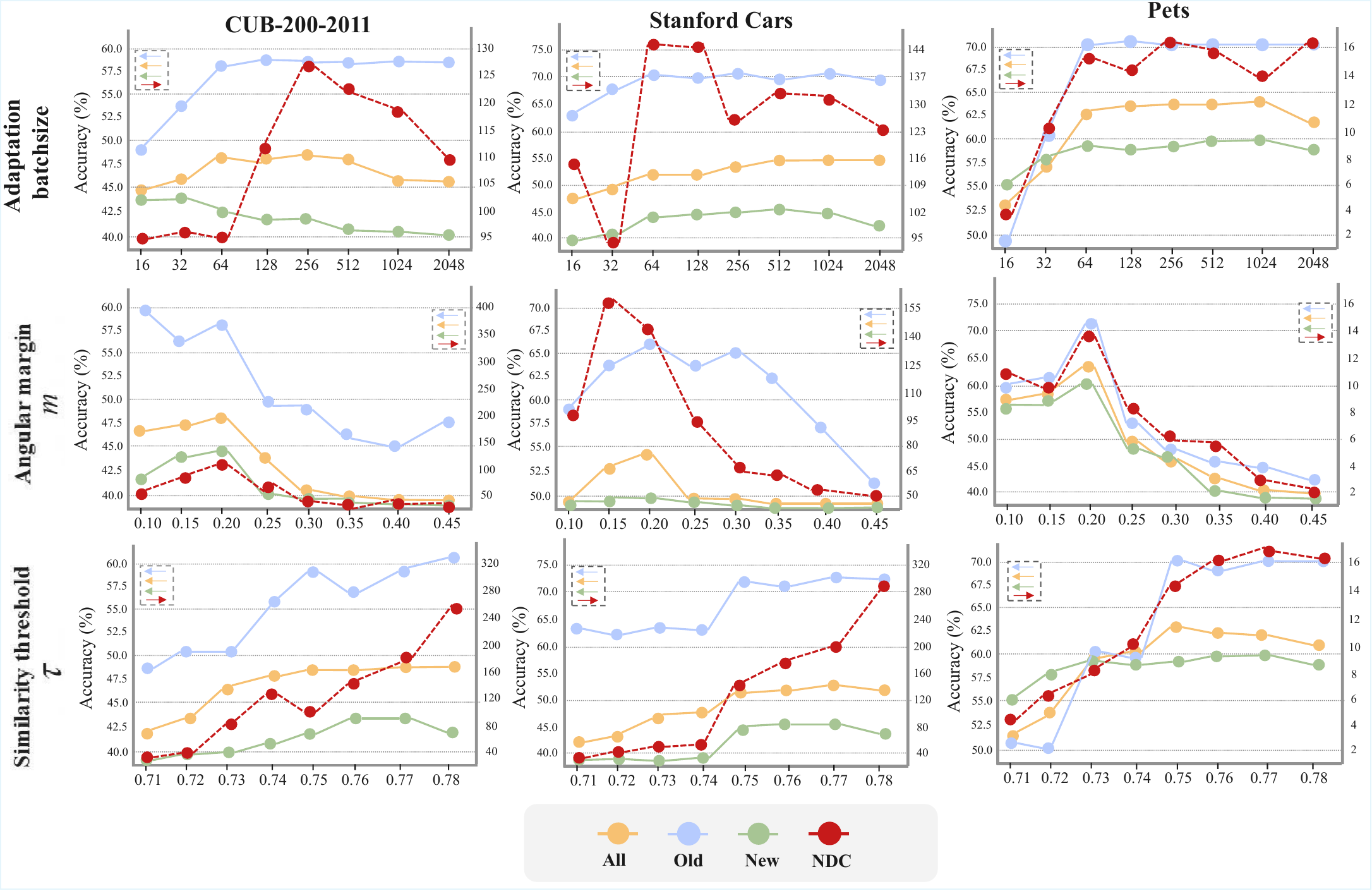}
    \caption{\textbf{Hyperparameter analysis on CUB-200-2011, Stanford Cars, and Oxford Pets datasets.} Each column corresponds to one dataset, showing the effects of adaptation batch size, angular margin $m$, and similarity threshold $\tau$ on accuracy (\textit{All}, \textit{Old}, \textit{New}) and the number of newly discovered categories (NDC).
}
    \label{fig:hyperparam}
    \vspace{-5mm}
\end{figure}

\vspace*{-0.4cm}
\paragraph{Impact of MLC on Feature Distributions.}
To further validate the effectiveness of the margin-aware logit calibration (MLC) module, 
we visualize the angular distributions of features and prototypes on the Pets dataset (see Fig.~\ref{fig:mlc_angle}). 
The left plot shows the angles between sample features and their class prototypes. 
Compared with the baseline (mean $64.55^\circ$), MLC reduces the average angle to $35.83^\circ$, indicating stronger intra-class compactness. 
The right plot presents the angles between different prototypes, where MLC enlarges the mean angle from $27.98^\circ$ to $74.15^\circ$, suggesting enhanced inter-class separability. These results confirm that MLC produces a more discriminative embedding space conducive to novel category discovery.

% These results confirm that MLC produces a more discriminative embedding space conducive to novel category discovery.

% \begin{figure}[h]
%     \centering
%     \begin{minipage}[t]{0.49\linewidth}
%         \centering
%         \includegraphics[width=\linewidth]{images/MLC-Inner.pdf}
%         % \caption*{(a) The Angle from the sample to the prototype.}
%     \end{minipage}
%     % \hfill
%     \begin{minipage}[t]{0.49\linewidth}
%         \centering
%         \includegraphics[width=\linewidth]{images/MLC-Outer.pdf}
%         % \caption*{(b) The Angle between prototypes.}
%     \end{minipage}
%     \caption{
% \textbf{Angular analysis of the margin-aware logit calibration (MLC) on the Oxford Pets dataset.} 
% Left: angles between samples and their class prototypes; Right: angles between different class prototypes.
% }
%     \label{fig:mlc_angle}
% \end{figure}

\vspace*{-0.5cm}
\paragraph{Hyperparameter analysis.} We analyze the sensitivity of TALON to three key hyperparameters: adaptation batch size, angular margin $m$, and similarity threshold $\tau$, as shown in Fig.~\ref{fig:hyperparam}. 
In each plot, the first three curves (blue, yellow, green) correspond to the left $y$-axis, while the red curve (NDC) refers to the right $y$-axis. 
Overall, performance improves with larger adaptation batches before saturating when redundancy appears. 
A moderate angular margin $m$ yields the best trade-off between intra-class compactness and inter-class separation, while an overly significant  margin harms feature geometry. 
As the similarity threshold $\tau$ increases, the model becomes more conservative, discovering more novel categories (higher NDC) with only minor impact on known-class accuracy.

% Overall, performance improves with larger adaptation batches before plateauing, and a moderate margin $m$ achieves the best balance between intra-class compactness and inter-class separation.
% As $\tau$ increases, the model becomes more conservative, discovering more novel categories (higher NDC) with minimal effect on known-class accuracy.
% Overall, TALON remains robust across a wide range of hyperparameter choices.

\section{Conclusion}
This paper presents a test-time adaptation framework for On-the-fly Category Discovery (OCD). To address the limitations of existing hash-based methods prone to information loss and category explosion, we propose a test-time knowledge accumulation strategy that combines semantic-aware prototype refinement with stable encoder adaptation. These components allow the model to dynamically absorb new information from streaming data while maintaining recognition stability. In addition, a Margin-aware Logit Calibration is introduced during offline training to enlarge inter-class margins and reserve embedding space for future category discovery. Experiments on standard OCD benchmarks show that our method consistently surpasses state-of-the-art approaches, significantly improving novel-class accuracy and mitigating category explosion.

\section{Acknowledgments}

The paper is supported in part by Beijing Smart Agriculture Innovation Consortium Project (BAIC10-2025). The authors gratefully acknowledge National Innovation Center for Digital Fishery - China Agricultural University, State Key Laboratory of Efficient Utilization of Agricultural Water Resources- China Agricultural University, Key Laboratory of Agricultural Informatization Standardization - MARA, P. R. China, Key Laboratory of Smart Farming Technologies for Aquatic Animals and Livestock - MARA, P. R. China, National Innovation Center for Digital Agricultural Products Circulation - MARA, P. R. China.
{
    \small
    \bibliographystyle{ieeenat_fullname}
    \bibliography{main}

\begin{thebibliography}{62}
\providecommand{\natexlab}[1]{#1}
\providecommand{\url}[1]{\texttt{#1}}
\expandafter\ifx\csname urlstyle\endcsname\relax
  \providecommand{\doi}[1]{doi: #1}\else
  \providecommand{\doi}{doi: \begingroup \urlstyle{rm}\Url}\fi

\bibitem[Bossard et~al.(2014)Bossard, Guillaumin, and Van~Gool]{food}
Lukas Bossard, Matthieu Guillaumin, and Luc Van~Gool.
\newblock Food-101--mining discriminative components with random forests.
\newblock In \emph{European Conference on Computer Vision}, pages 446--461. Springer, 2014.

\bibitem[Caron et~al.(2021)Caron, Touvron, Misra, J{\'e}gou, Mairal, Bojanowski, and Joulin]{caron2021emerging}
Mathilde Caron, Hugo Touvron, Ishan Misra, Herv{\'e} J{\'e}gou, Julien Mairal, Piotr Bojanowski, and Armand Joulin.
\newblock Emerging properties in self-supervised vision transformers.
\newblock In \emph{Proceedings of the IEEE/CVF International Conference on Computer Vision}, pages 9650--9660, 2021.

\bibitem[Chen et~al.(2021)Chen, Li, Bai, Yang, Jiang, and Miao]{chen2021review}
Leiyu Chen, Shaobo Li, Qiang Bai, Jing Yang, Sanlong Jiang, and Yanming Miao.
\newblock Review of image classification algorithms based on convolutional neural networks.
\newblock \emph{Remote Sensing}, 13\penalty0 (22):\penalty0 4712, 2021.

\bibitem[Chi et~al.(2022)Chi, Gu, Liu, Wang, Yu, and Tang]{chi2022metafscil}
Zhixiang Chi, Li Gu, Huan Liu, Yang Wang, Yuanhao Yu, and Jin Tang.
\newblock Metafscil: A meta-learning approach for few-shot class incremental learning.
\newblock In \emph{Proceedings of the IEEE/CVF Conference on Computer Vision and Pattern Recognition}, pages 14166--14175, 2022.

\bibitem[Chi et~al.(2024)Chi, Gu, Zhong, Liu, YU, Plataniotis, and Wang]{chiadapting}
Zhixiang Chi, Li Gu, Tao Zhong, Huan Liu, YUANHAO YU, Konstantinos~N Plataniotis, and Yang Wang.
\newblock Adapting to distribution shift by visual domain prompt generation.
\newblock In \emph{The Twelfth International Conference on Learning Representations}, 2024.

\bibitem[Chi et~al.(2025{\natexlab{a}})Chi, Gu, Liu, Wang, Wu, Wang, and Plataniotis]{chilearning}
Zhixiang Chi, Li Gu, Huan Liu, Ziqiang Wang, Yanan Wu, Yang Wang, and Konstantinos~N Plataniotis.
\newblock Learning to adapt frozen clip for few-shot test-time domain adaptation.
\newblock In \emph{The Thirteenth International Conference on Learning Representations}, 2025{\natexlab{a}}.

\bibitem[Chi et~al.(2025{\natexlab{b}})Chi, Wu, Gu, Liu, Wang, Zhang, Wang, and Plataniotis]{chi2025plug}
Zhixiang Chi, Yanan Wu, Li Gu, Huan Liu, Ziqiang Wang, Yang Zhang, Yang Wang, and Konstantinos Plataniotis.
\newblock Plug-in feedback self-adaptive attention in clip for training-free open-vocabulary segmentation.
\newblock In \emph{Proceedings of the IEEE/CVF International Conference on Computer Vision}, pages 22815--22825, 2025{\natexlab{b}}.

\bibitem[Choi et~al.(2024)Choi, Kang, and Cho]{choi2024contrastive}
Sua Choi, Dahyun Kang, and Minsu Cho.
\newblock Contrastive mean-shift learning for generalized category discovery.
\newblock In \emph{Proceedings of the IEEE/CVF Conference on Computer Vision and Pattern Recognition}, pages 23094--23104, 2024.

\bibitem[Dong et~al.(2025)Dong, Chatzi, and Fink]{dong2025towards}
Hao Dong, Eleni Chatzi, and Olga Fink.
\newblock Towards robust multimodal open-set test-time adaptation via adaptive entropy-aware optimization.
\newblock In \emph{International Conference on Learning Representations}, pages 41799--41824, 2025.

\bibitem[Dosovitskiy(2021)]{dosovitskiy2020image}
Alexey Dosovitskiy.
\newblock An image is worth 16x16 words: Transformers for image recognition at scale.
\newblock In \emph{International Conference on Learning Representations}, pages 611--631, 2021.

\bibitem[Du et~al.(2023)Du, Chang, Liang, Hospedales, Song, and Ma]{SMILE}
Ruoyi Du, Dongliang Chang, Kongming Liang, Timothy Hospedales, Yi-Zhe Song, and Zhanyu Ma.
\newblock On-the-fly category discovery.
\newblock In \emph{Proceedings of the IEEE/CVF Conference on Computer Vision and Pattern Recognition}, pages 11691--11700, 2023.

\bibitem[Fini et~al.(2021)Fini, Sangineto, Lathuili{\`e}re, Zhong, Nabi, and Ricci]{fini2021unified}
Enrico Fini, Enver Sangineto, St{\'e}phane Lathuili{\`e}re, Zhun Zhong, Moin Nabi, and Elisa Ricci.
\newblock A unified objective for novel class discovery.
\newblock In \emph{Proceedings of the IEEE/CVF International Conference on Computer Vision}, pages 9284--9292, 2021.

\bibitem[Gandelsman et~al.(2022)Gandelsman, Sun, Chen, and Efros]{gandelsman2022test}
Yossi Gandelsman, Yu Sun, Xinlei Chen, and Alexei Efros.
\newblock Test-time training with masked autoencoders.
\newblock \emph{Advances in Neural Information Processing Systems}, 35:\penalty0 29374--29385, 2022.

\bibitem[Gong et~al.(2022)Gong, Jeong, Kim, Kim, Shin, and Lee]{gong2022note}
Taesik Gong, Jongheon Jeong, Taewon Kim, Yewon Kim, Jinwoo Shin, and Sung-Ju Lee.
\newblock Note: Robust continual test-time adaptation against temporal correlation.
\newblock \emph{Advances in Neural Information Processing Systems}, 35:\penalty0 27253--27266, 2022.

\bibitem[Han et~al.(2019)Han, Vedaldi, and Zisserman]{NCD}
Kai Han, Andrea Vedaldi, and Andrew Zisserman.
\newblock Learning to discover novel visual categories via deep transfer clustering.
\newblock In \emph{Proceedings of the IEEE/CVF International Conference on Computer Vision}, pages 8401--8409, 2019.

\bibitem[Han et~al.(2021)Han, Rebuffi, Ehrhardt, Vedaldi, and Zisserman]{RankStat}
Kai Han, Sylvestre-Alvise Rebuffi, Sebastien Ehrhardt, Andrea Vedaldi, and Andrew Zisserman.
\newblock Autonovel: Automatically discovering and learning novel visual categories.
\newblock \emph{IEEE Transactions on Pattern Analysis and Machine Intelligence}, 44\penalty0 (10):\penalty0 6767--6781, 2021.

\bibitem[Hartigan(1975)]{hartigan1975clustering}
John~A Hartigan.
\newblock \emph{Clustering algorithms}.
\newblock John Wiley \& Sons, Inc., 1975.

\bibitem[He et~al.(2016)He, Zhang, Ren, and Sun]{he2016deep}
Kaiming He, Xiangyu Zhang, Shaoqing Ren, and Jian Sun.
\newblock Deep residual learning for image recognition.
\newblock In \emph{Proceedings of the IEEE/CVF Conference on Computer Vision and Pattern Recognition}, pages 770--778, 2016.

\bibitem[He et~al.(2022)He, Chen, Xie, Li, Doll{\'a}r, and Girshick]{he2022masked}
Kaiming He, Xinlei Chen, Saining Xie, Yanghao Li, Piotr Doll{\'a}r, and Ross Girshick.
\newblock Masked autoencoders are scalable vision learners.
\newblock In \emph{Proceedings of the IEEE/CVF Conference on Computer Vision and Pattern Recognition}, pages 16000--16009, 2022.

\bibitem[Hong et~al.(2021)Hong, Han, Yao, Gao, Zhang, Plaza, and Chanussot]{hong2021spectralformer}
Danfeng Hong, Zhu Han, Jing Yao, Lianru Gao, Bing Zhang, Antonio Plaza, and Jocelyn Chanussot.
\newblock Spectralformer: Rethinking hyperspectral image classification with transformers.
\newblock \emph{IEEE Transactions on Geoscience and Remote Sensing}, 60:\penalty0 1--15, 2021.

\bibitem[Jia et~al.(2021)Jia, Han, Zhu, and Green]{WTAjia2021joint}
Xuhui Jia, Kai Han, Yukun Zhu, and Bradley Green.
\newblock Joint representation learning and novel category discovery on single-and multi-modal data.
\newblock In \emph{Proceedings of the IEEE/CVF International Conference on Computer Vision}, pages 610--619, 2021.

\bibitem[Jung and Wang(2025)]{NCD-DLT}
Hoin Jung and Xiaoqian Wang.
\newblock Towards on-the-fly novel category discovery in dynamic long-tailed distributions.
\newblock In \emph{IEEE/CVF Winter Conference on Applications of Computer Vision}, 2025.

\bibitem[Khosla et~al.(2020)Khosla, Teterwak, Wang, Sarna, Tian, Isola, Maschinot, Liu, and Krishnan]{khosla2020supervised}
Prannay Khosla, Piotr Teterwak, Chen Wang, Aaron Sarna, Yonglong Tian, Phillip Isola, Aaron Maschinot, Ce Liu, and Dilip Krishnan.
\newblock Supervised contrastive learning.
\newblock \emph{Advances in neural information processing systems}, 33:\penalty0 18661--18673, 2020.

\bibitem[Krause et~al.(2013)Krause, Stark, Deng, and Fei-Fei]{scars}
Jonathan Krause, Michael Stark, Jia Deng, and Li Fei-Fei.
\newblock 3d object representations for fine-grained categorization.
\newblock In \emph{Proceedings of the IEEE international conference on computer vision workshops}, pages 554--561, 2013.

\bibitem[Krizhevsky et~al.(2009)Krizhevsky, Hinton, et~al.]{cifar}
Alex Krizhevsky, Geoffrey Hinton, et~al.
\newblock Learning multiple layers of features from tiny images.
\newblock 2009.

\bibitem[Li et~al.(2018)Li, Yang, Song, and Hospedales]{MLDGli2018learning}
Da Li, Yongxin Yang, Yi-Zhe Song, and Timothy Hospedales.
\newblock Learning to generalize: Meta-learning for domain generalization.
\newblock In \emph{Proceedings of the AAAI conference on artificial intelligence}, 2018.

\bibitem[Li et~al.(2023{\natexlab{a}})Li, Fan, Huo, and Gao]{li2023modeling}
Wenbin Li, Zhichen Fan, Jing Huo, and Yang Gao.
\newblock Modeling inter-class and intra-class constraints in novel class discovery.
\newblock In \emph{Proceedings of the IEEE/CVF Conference on Computer Vision and Pattern Recognition}, pages 3449--3458, 2023{\natexlab{a}}.

\bibitem[Li et~al.(2025)Li, Fang, and Li]{li2025generalized}
Xiao Li, Min Fang, and HaiXiang Li.
\newblock Generalized category discovery with unknown sample generation.
\newblock \emph{IEEE Transactions on Image Processing}, 2025.

\bibitem[Li et~al.(2023{\natexlab{b}})Li, Xu, Su, and Jia]{li2023robustness}
Yushu Li, Xun Xu, Yongyi Su, and Kui Jia.
\newblock On the robustness of open-world test-time training: Self-training with dynamic prototype expansion.
\newblock In \emph{Proceedings of the IEEE/CVF International Conference on Computer Vision}, pages 11836--11846, 2023{\natexlab{b}}.

\bibitem[Lim et~al.(2023)Lim, Kim, Choo, and Choi]{lim2023ttn}
Hyesu Lim, Byeonggeun Kim, Jaegul Choo, and Sungha Choi.
\newblock Ttn: A domain-shift aware batch normalization in test-time adaptation.
\newblock In \emph{International Conference on Learning Representations}, pages 27752--27770, 2023.

\bibitem[Liu et~al.(2022)Liu, Gu, Chi, Wang, Yu, Chen, and Tang]{liu2022few}
Huan Liu, Li Gu, Zhixiang Chi, Yang Wang, Yuanhao Yu, Jun Chen, and Jin Tang.
\newblock Few-shot class-incremental learning via entropy-regularized data-free replay.
\newblock In \emph{European Conference on Computer Vision}, pages 146--162. Springer, 2022.

\bibitem[Liu et~al.(2023)Liu, Chi, Yu, Wang, Chen, and Tang]{liu2023meta}
Huan Liu, Zhixiang Chi, Yuanhao Yu, Yang Wang, Jun Chen, and Jin Tang.
\newblock Meta-auxiliary learning for future depth prediction in videos.
\newblock In \emph{Proceedings of the IEEE/CVF Winter Conference on Applications of Computer Vision}, pages 5756--5765, 2023.

\bibitem[Liu et~al.(2025)Liu, Pu, Zheng, Li, Sebe, and Zhong]{DiffGRE}
Xiao Liu, Nan Pu, Haiyang Zheng, Wenjing Li, Nicu Sebe, and Zhun Zhong.
\newblock Generate, refine, and encode: Leveraging synthesized novel samples for on-the-fly fine-grained category discovery.
\newblock In \emph{Proceedings of the IEEE/CVF International Conference on Computer Vision}, pages 1078--1087, 2025.

\bibitem[Ma et~al.(2024{\natexlab{a}})Ma, Zhu, Zhong, Liu, Zhang, and Liu]{ma2024happy}
Shijie Ma, Fei Zhu, Zhun Zhong, Wenzhuo Liu, Xu-Yao Zhang, and Cheng-Lin Liu.
\newblock Happy: A debiased learning framework for continual generalized category discovery.
\newblock \emph{Advances in Neural Information Processing Systems}, 37:\penalty0 50850--50875, 2024{\natexlab{a}}.

\bibitem[Ma et~al.(2024{\natexlab{b}})Ma, Zhu, Zhong, Zhang, and Liu]{ma2024active}
Shijie Ma, Fei Zhu, Zhun Zhong, Xu-Yao Zhang, and Cheng-Lin Liu.
\newblock Active generalized category discovery.
\newblock In \emph{Proceedings of the IEEE/CVF Conference on Computer Vision and Pattern Recognition}, pages 16890--16900, 2024{\natexlab{b}}.

\bibitem[Ma et~al.(2025{\natexlab{a}})Ma, Ge, Wang, Guo, Ge, and Shan]{ma2025genhancer}
Shijie Ma, Yuying Ge, Teng Wang, Yuxin Guo, Yixiao Ge, and Ying Shan.
\newblock Genhancer: Imperfect generative models are secretly strong vision-centric enhancers.
\newblock In \emph{Proceedings of the IEEE/CVF International Conference on Computer Vision}, pages 24402--24412, 2025{\natexlab{a}}.

\bibitem[Ma et~al.(2025{\natexlab{b}})Ma, Zhu, Zhang, and Liu]{ma2025protogcd}
Shijie Ma, Fei Zhu, Xu-Yao Zhang, and Cheng-Lin Liu.
\newblock Protogcd: Unified and unbiased prototype learning for generalized category discovery.
\newblock \emph{IEEE Transactions on Pattern Analysis and Machine Intelligence}, 2025{\natexlab{b}}.

\bibitem[Mao et~al.(2023)Mao, Mohri, and Zhong]{mao2023cross}
Anqi Mao, Mehryar Mohri, and Yutao Zhong.
\newblock Cross-entropy loss functions: Theoretical analysis and applications.
\newblock In \emph{International conference on machine learning}, pages 23803--23828. pmlr, 2023.

\bibitem[Parkhi et~al.(2012)Parkhi, Vedaldi, Zisserman, and Jawahar]{pets}
Omkar~M Parkhi, Andrea Vedaldi, Andrew Zisserman, and CV Jawahar.
\newblock Cats and dogs.
\newblock In \emph{2012 IEEE conference on computer vision and pattern recognition}, pages 3498--3505. IEEE, 2012.

\bibitem[Pu et~al.(2023)Pu, Zhong, and Sebe]{pu2023dynamic}
Nan Pu, Zhun Zhong, and Nicu Sebe.
\newblock Dynamic conceptional contrastive learning for generalized category discovery.
\newblock In \emph{Proceedings of the IEEE/CVF Conference on Computer Vision and Pattern Recognition}, pages 7579--7588, 2023.

\bibitem[Radford et~al.(2021)Radford, Kim, Hallacy, Ramesh, Goh, Agarwal, Sastry, Askell, Mishkin, Clark, et~al.]{radford2021learning}
Alec Radford, Jong~Wook Kim, Chris Hallacy, Aditya Ramesh, Gabriel Goh, Sandhini Agarwal, Girish Sastry, Amanda Askell, Pamela Mishkin, Jack Clark, et~al.
\newblock Learning transferable visual models from natural language supervision.
\newblock In \emph{International conference on machine learning}, pages 8748--8763, 2021.

\bibitem[Rathore et~al.(2025)Rathore, Dutta, Mehrotra, Kira, Banerjee, et~al.]{rathore2025domain}
Vaibhav Rathore, Saikat Dutta, Sarthak Mehrotra, Zsolt Kira, Biplab Banerjee, et~al.
\newblock When domain generalization meets generalized category discovery: An adaptive task-arithmetic driven approach.
\newblock In \emph{Proceedings of the IEEE/CVF Conference on Computer Vision and Pattern Recognition}, pages 4905--4915, 2025.

\bibitem[Rongali et~al.(2024)Rongali, Mehrotra, Jha, Bose, Gupta, Singha, Banerjee, et~al.]{rongali2024cdad}
Sai~Bhargav Rongali, Sarthak Mehrotra, Ankit Jha, Shirsha Bose, Tanisha Gupta, Mainak Singha, Biplab Banerjee, et~al.
\newblock Cdad-net: Bridging domain gaps in generalized category discovery.
\newblock In \emph{Proceedings of the IEEE/CVF Conference on Computer Vision and Pattern Recognition}, pages 2616--2626, 2024.

\bibitem[Roy et~al.(2022)Roy, Liu, Zhong, Sebe, and Ricci]{roy2022class}
Subhankar Roy, Mingxuan Liu, Zhun Zhong, Nicu Sebe, and Elisa Ricci.
\newblock Class-incremental novel class discovery.
\newblock In \emph{European Conference on Computer Vision}, pages 317--333, 2022.

\bibitem[Russakovsky et~al.(2015)Russakovsky, Deng, Su, Krause, Satheesh, Ma, Huang, Karpathy, Khosla, Bernstein, et~al.]{2015imagenet}
Olga Russakovsky, Jia Deng, Hao Su, Jonathan Krause, Sanjeev Satheesh, Sean Ma, Zhiheng Huang, Andrej Karpathy, Aditya Khosla, Michael Bernstein, et~al.
\newblock Imagenet large scale visual recognition challenge.
\newblock \emph{International journal of computer vision}, 115\penalty0 (3):\penalty0 211--252, 2015.

\bibitem[Schneider et~al.(2020)Schneider, Rusak, Eck, Bringmann, Brendel, and Bethge]{schneider2020improving}
Steffen Schneider, Evgenia Rusak, Luisa Eck, Oliver Bringmann, Wieland Brendel, and Matthias Bethge.
\newblock Improving robustness against common corruptions by covariate shift adaptation.
\newblock \emph{Advances in neural information processing systems}, 33:\penalty0 11539--11551, 2020.

\bibitem[Sun et~al.(2020)Sun, Wang, Liu, Miller, Efros, and Hardt]{sun2020test}
Yu Sun, Xiaolong Wang, Zhuang Liu, John Miller, Alexei Efros, and Moritz Hardt.
\newblock Test-time training with self-supervision for generalization under distribution shifts.
\newblock In \emph{International conference on machine learning}, pages 9229--9248, 2020.

\bibitem[Vaze et~al.(2022)Vaze, Han, Vedaldi, and Zisserman]{vaze2022generalized}
Sagar Vaze, Kai Han, Andrea Vedaldi, and Andrew Zisserman.
\newblock Generalized category discovery.
\newblock In \emph{Proceedings of the IEEE/CVF Conference on Computer Vision and Pattern Recognition}, pages 7492--7501, 2022.

\bibitem[Wah et~al.(2011)Wah, Branson, Welinder, Perona, and Belongie]{CUB}
Catherine Wah, Steve Branson, Peter Welinder, Pietro Perona, and Serge Belongie.
\newblock The caltech-ucsd birds-200-2011 dataset.
\newblock 2011.

\bibitem[Wang et~al.(2021)Wang, Shelhamer, Liu, Olshausen, and Darrell]{wang2020tent}
Dequan Wang, Evan Shelhamer, Shaoteng Liu, Bruno Olshausen, and Trevor Darrell.
\newblock Tent: Fully test-time adaptation by entropy minimization.
\newblock In \emph{International Conference on Learning Representations}, pages 2928--2942, 2021.

\bibitem[Wang et~al.(2025)Wang, Vaze, and Han]{wang2024hilo}
Hongjun Wang, Sagar Vaze, and Kai Han.
\newblock Hilo: A learning framework for generalized category discovery robust to domain shifts.
\newblock In \emph{International Conference on Learning Representations}, pages 3773--3814, 2025.

\bibitem[Wang et~al.(2024)Wang, Chi, Wu, Gu, Liu, Plataniotis, and Wang]{wang2024distribution}
Ziqiang Wang, Zhixiang Chi, Yanan Wu, Li Gu, Zhi Liu, Konstantinos Plataniotis, and Yang Wang.
\newblock Distribution alignment for fully test-time adaptation with dynamic online data streams.
\newblock In \emph{European Conference on Computer Vision}, pages 332--349. Springer, 2024.

\bibitem[Wu et~al.(2023)Wu, Chi, Wang, and Feng]{wu2023metagcd}
Yanan Wu, Zhixiang Chi, Yang Wang, and Songhe Feng.
\newblock Metagcd: Learning to continually learn in generalized category discovery.
\newblock In \emph{Proceedings of the IEEE/CVF International Conference on Computer Vision}, pages 1655--1665, 2023.

\bibitem[Wu et~al.(2024)Wu, Chi, Wang, Plataniotis, and Feng]{wu2024test}
Yanan Wu, Zhixiang Chi, Yang Wang, Konstantinos~N Plataniotis, and Songhe Feng.
\newblock Test-time domain adaptation by learning domain-aware batch normalization.
\newblock In \emph{Proceedings of the AAAI Conference on Artificial Intelligence}, pages 15961--15969, 2024.

\bibitem[Yang et~al.(2023)Yang, Wang, Deng, and Zhang]{yang2023bootstrap}
Muli Yang, Liancheng Wang, Cheng Deng, and Hanwang Zhang.
\newblock Bootstrap your own prior: Towards distribution-agnostic novel class discovery.
\newblock In \emph{Proceedings of the IEEE/CVF Conference on Computer Vision and Pattern Recognition}, pages 3459--3468, 2023.

\bibitem[You et~al.(2021)You, Li, and Zhao]{you2021test}
Fuming You, Jingjing Li, and Zhou Zhao.
\newblock Test-time batch statistics calibration for covariate shift.
\newblock \emph{arXiv preprint arXiv:2110.04065}, 2021.

\bibitem[Yuan et~al.(2023)Yuan, Xie, and Li]{yuan2023robust}
Longhui Yuan, Binhui Xie, and Shuang Li.
\newblock Robust test-time adaptation in dynamic scenarios.
\newblock In \emph{Proceedings of the IEEE/CVF Conference on Computer Vision and Pattern Recognition}, pages 15922--15932, 2023.

\bibitem[Zhang et~al.(2022)Zhang, Levine, and Finn]{zhang2022memo}
Marvin Zhang, Sergey Levine, and Chelsea Finn.
\newblock Memo: Test time robustness via adaptation and augmentation.
\newblock \emph{Advances in neural information processing systems}, 35:\penalty0 38629--38642, 2022.

\bibitem[Zhang et~al.(2025)Zhang, Jiang, Feng, Wu, Zhao, Wan, Mingqian, Jin, and Gao]{GM}
Xinwei Zhang, Jianwen Jiang, Yutong Feng, Zhi-Fan Wu, Xibin Zhao, Hai Wan, Tang Mingqian, Rong Jin, and Yue Gao.
\newblock Grow and merge: A unified framework for continuous categories discovery.
\newblock In \emph{Conference on Neural Information Processing Systems}, 2025.

\bibitem[Zhao and Mac~Aodha(2023)]{zhao2023incremental}
Bingchen Zhao and Oisin Mac~Aodha.
\newblock Incremental generalized category discovery.
\newblock In \emph{Proceedings of the IEEE/CVF International Conference on Computer Vision}, pages 19137--19147, 2023.

\bibitem[Zheng et~al.(2024)Zheng, Pu, Li, Sebe, and Zhong]{PHE}
Haiyang Zheng, Nan Pu, Wenjing Li, Nicu Sebe, and Zhun Zhong.
\newblock Prototypical hash encoding for on-the-fly fine-grained category discovery.
\newblock \emph{Advances in Neural Information Processing Systems}, 37:\penalty0 101428--101455, 2024.

\bibitem[Zhong et~al.(2021)Zhong, Fini, Roy, Luo, Ricci, and Sebe]{zhong2021neighborhood}
Zhun Zhong, Enrico Fini, Subhankar Roy, Zhiming Luo, Elisa Ricci, and Nicu Sebe.
\newblock Neighborhood contrastive learning for novel class discovery.
\newblock In \emph{Proceedings of the IEEE/CVF Conference on Computer Vision and Pattern Recognition}, pages 10867--10875, 2021.

\end{thebibliography}
}

\appendix

\setcounter{page}{1}
\maketitlesupplementary

\section{Implementation Details}
% \vspace{-1mm}
\subsection{Datasets Details and Evaluation Metric Details}
% \vspace{-2mm}
\textbf{Dataset Details.}  To comprehensively evaluate our framework across different levels of semantic complexity, we adopt a diverse set of benchmark datasets that cover both coarse-grained and fine-grained recognition scenarios.As shown in Tab.~\ref{tab:table_coarse}, the coarse-grained datasets used in our experiments include CIFAR-10 \cite{cifar}, CIFAR-100 \cite{cifar}, and ImageNet-100 \cite{2015imagenet}.These datasets exhibit large inter-class variation and broad semantic categories, making them suitable for evaluating category discovery under generic object recognition settings. In contrast, the fine-grained datasets summarized in Tab.~\ref{tab:table_fine}. CUB-200-2011 \cite{CUB}, Stanford Cars \cite{scars},Oxford Pets \cite{pets}, and Food-101 \cite{food} contain subtle visual distinctions among classes and therefore provide a more challenging testing ground for fine-grained category discovery. By jointly employing datasets of different granularity, our evaluation ensures that the proposed method is rigorously tested under diverse and realistic visual recognition conditions.Specifically, 50\% of the samples from the seen categories are used to form the labeled training set $\mathcal{D}_S$, while the remainder forms the unlabeled set $\mathcal{D}_Q$ for on-the-fly testing.

% ----------------------------------------------------
% Coarse-grained table
% ----------------------------------------------------
% \vspace{-1mm}
\begin{table}[H]
\vspace{-3mm}
\centering
\caption{Statistics of coarse-grained datasets.}
\vspace{-3mm}
\label{tab:table_coarse}
\small
\begin{tabular}{lccc}
\toprule
Dataset & CIFAR10 & CIFAR100 & ImageNet-100  \\
\midrule
$|Y_Q|$ & 5 & 80 & 80  \\
$|Y_S|$ & 10 & 100 & 100 \\
\midrule
$|D_S|$ & 12.5K & 20.0K & 31.9K   \\
$|D_Q|$ & 37.5K & 30.0K & 95.3K   \\
\bottomrule
\end{tabular}
\end{table}

\vspace{-3mm}
\begin{table}[H]
\vspace{-3mm}
\centering
\caption{Statistics of fine-grained datasets.}
\vspace{-3mm}
\label{tab:table_fine}
\resizebox{\linewidth}{!}{
\begin{tabular}{lcccc}
\toprule
Dataset & CUB-200-2011 & Stanford Cars & Oxford Pets& Food101 \\
\midrule
$|Y_Q|$ & 100 & 98 & 19& 51 \\
$|Y_S|$ & 200 & 196 & 38 & 101\\
\midrule
$|D_S|$ & 1.5K & 2.0K & 0.9K& 19.1K \\
$|D_Q|$ & 4.5K & 6.1K & 2.7K& 56.6K \\
\bottomrule
\end{tabular}}
\end{table}

\noindent\textbf{Evaluation Metric Details.} Following \cite{SMILE}, we adopt two protocols for evaluation termed \textit{Greedy-Hungarian} and \textit{Strict-Hungarian} for comprehensive comparisons, where their difference is clearly illustrated in \cite{vaze2022generalized}.
During testing, samples sharing the same category descriptor form a cluster, and only the top-$|\mathcal{Y}_Q|$ clusters by size are retained, with the rest treated as misclassified.
For \textit{Greedy-Hungarian}, accuracy is computed separately on the “Old” and “New” subsets, providing independent evaluations of known and novel classes.
In contrast, \textit{Strict-Hungarian} calculates accuracy over the entire query set, avoiding repeated cluster assignments between subsets.
The overall accuracy is obtained via the Hungarian matching:
\vspace{-1mm}
 \begin{equation}
ACC=\max\limits_{p\in \mathcal{P}(\mathcal{Y}_Q)}\frac{1}{ \left| \mathcal{D}_Q \right| }\sum^{\left| \mathcal{D}_Q \right|}_{i=1}\mathds{1}[y_i = p(\hat{y}_i)],
\end{equation}
% \vspace{-1mm}
where $y_i$ is the ground truth label, $\hat{y}_i$ is the predicted label decided by cluster indices, and $\mathcal{P}(\mathcal{Y}_Q) $ is the set of all permutations of ground truth lables.
\subsection{Training Details}
We adopt two vision backbones in our experiments, namely CLIP ViT-B/16 and DINOv2-ViT-Base, and build our framework on top of their pre-trained weights.  For CLIP, we use the publicly released ViT-B/16 model, who outputs 512-dimensional image embeddings. For DINOv2, we use the official DINOv2-ViT-Base checkpoint,  its output features are further projected into the 768-dimensional visual space.

We train the model using the AdamW optimizer with a weight decay of 0.05 and a cosine learning rate scheduler with a minimum learning rate of $1\mathrm{e}{-5}$. Following our implementation, we employ parameter-specific learning rates: the prototype layer is optimized with a learning rate of $1\mathrm{e}{-3}$, the last visual layer of the backbone is updated with a smaller learning rate of $1\mathrm{e}{-4}$, and the remaining parameters follow the base learning rate governed by the scheduler. The batch size is uniformly set to 128 for all datasets, and we train for 100 epochs. All experiments are conducted on NVIDIA RTX~3090 GPUs with 24GB memory, and we fix the random seed to 1028 to improve reproducibility.

\subsection{Compared Methods Details}
\textbf{Ranking Statistics (RankStat).}~\cite{RankStat}
AutoNovel adopts Ranking Statistics to characterize sample relationships, where the top-3 indices of feature embeddings are taken as category indicators. This formulation fits well within the On-the-Fly Category Discovery (OCD) paradigm and presents a strong benchmark for evaluating hash-based descriptors. For fairness, we reimplement Ranking Statistics using the same backbone (DINO-ViT-B-16) and preserve only the fully supervised training stage, since OCD does not allow any external data. The embedding dimension is fixed to 32, yielding a prediction space of $C_{32}^3=4,906$, which is comparable to our approach and SMILE, where a hash length of $L=12$ produces $2^{12}=4,096$ possible codes.

 \textbf{Winner-take-all (WTA).}~\cite{WTAjia2021joint}
To mitigate the bias of Ranking Statistics toward overly prominent features, the Winner-take-all (WTA) hashing strategy was proposed. Instead of relying on a global ranking of feature activations, WTA constructs descriptors by identifying the maximum index within each of several partitioned feature groups. Using a 48-dimensional embedding split into three groups, WTA generates $16^3=4096$ distinct codes, ensuring direct comparability with the other hashing-based baselines.

 \textbf{Sequential Leader Clustering (SLC).}~\cite{hartigan1975clustering}
For SLC, we adopt the same backbone and apply conventional supervised learning on the support set. During on-the-fly evaluation, SLC assigns labels using features extracted from the query samples. Hyperparameters are tuned on the CUB dataset and subsequently applied unchanged to all other datasets to maintain consistency and fairness across comparisons.

 \textbf{MLDG.}~\cite{MLDGli2018learning}
Different from the standard NCD setting, which leverages both support and query sets to learn discriminative features, OCD is closer to a domain-generalization problem: the model is trained on seen categories and must generalize to unseen ones at inference time. Therefore, we include MLDG~\cite{MLDGli2018learning}, a model-agnostic domain generalization algorithm, as a strong competitor. During training, samples from different classes are partitioned into meta-train and meta-test domains at each iteration to encourage generalizable representations.

For \textbf{SMILE}\cite{SMILE}, \textbf{PHE}\cite{PHE} and \textbf{DiffGRE}\cite{DiffGRE}, the best configuration of their original paper reports is adopted

\section{Additional Experiment and Analysis}
\subsection{Hyperparameter analysis.}

\begin{figure}[t]
    \centering
    \includegraphics[width=1\linewidth]{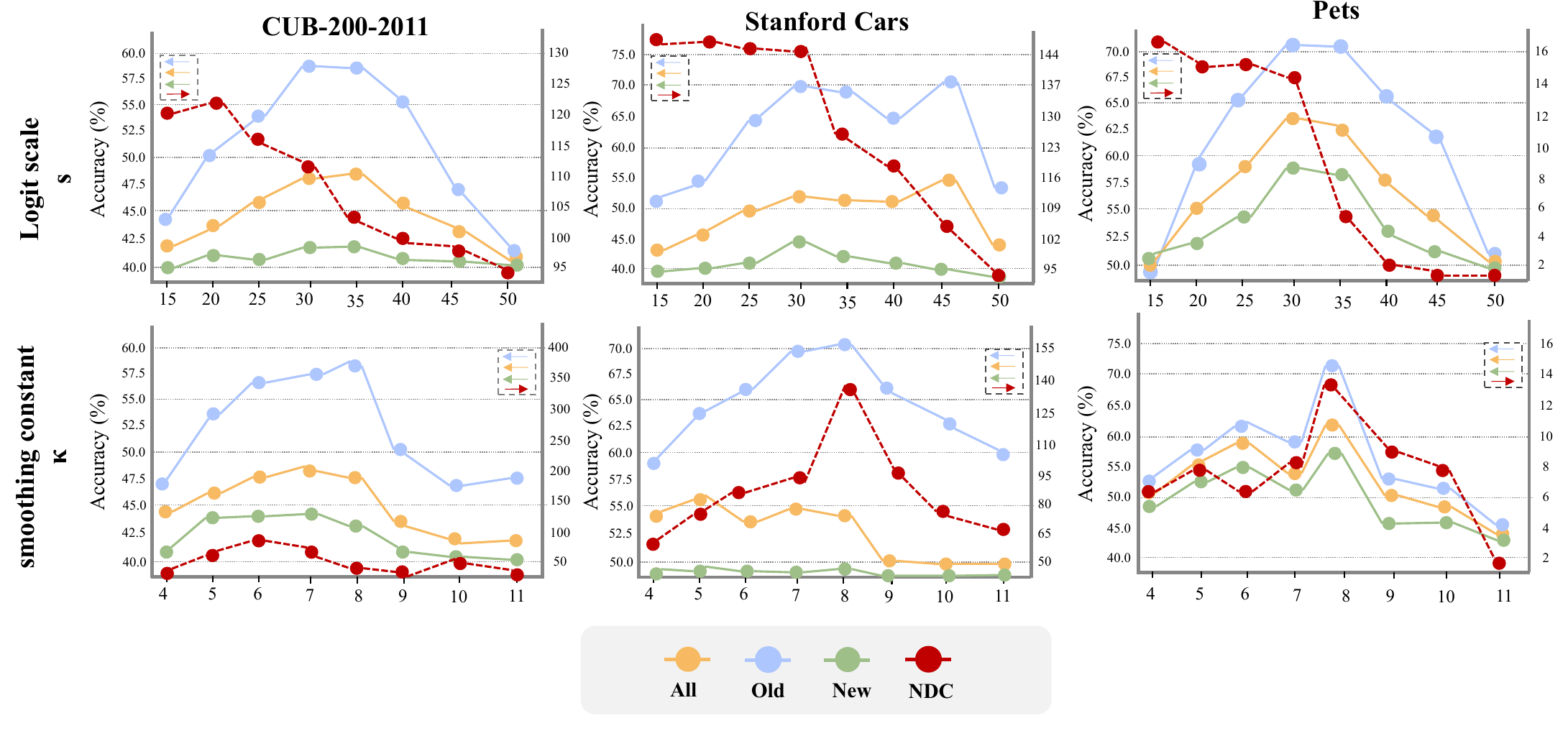}
    \caption{\textbf{Hyperparameter analysis on CUB-200-2011, Stanford Cars, and Oxford Pets datasets.} Each column corresponds to one dataset, showing the effects of  logit scale $s$ and the smoothing constant $\kappa$ on accuracy (\textit{All}, \textit{Old}, \textit{New}) and the number of newly discovered categories (NDC).
}
    \label{fig:hyperparam2}
    \vspace{-5mm}
\end{figure}

We further analyze the sensitivity of our method to two key hyper-parameters: the logit scale $s$ and the smoothing constant $\kappa$. The $s$ controls the magnitude of the logits in MLC, thereby affecting the margin strength and confidence of the classifier, while $\kappa$ appears in the Semantic-aware prototype update and balances how fast the prototypes are allowed to adapt to incoming test samples. To assess the robustness of our method, we conduct a series of experiments by varying $s$ in $\{10, 15, 20, 25, 30, 35, 40, 45, 50\}$ and $\kappa$  in $\{4, 5, 6, 7, 8, 9, 10, 11, 12\}$, and report the corresponding results in Fig.~\ref{fig:hyperparam2}.

\begin{table*}[ht]
\centering
\small
\setlength{\tabcolsep}{5pt}
\renewcommand{\arraystretch}{1.10}

\newcolumntype{C}[1]{>{\centering\arraybackslash}m{#1}}

\caption{\textbf{Comparison with State-of-the-Art Methods Using Different Backbones} }
\resizebox{\textwidth}{!}{
\begin{tabular}{C{1mm} C{24mm} C{16mm}
|ccc|ccc|ccc|ccc|ccc|ccc|ccc}
\toprule
\multirow{2}{*}{} 
& \multirow{2}{*}{\textbf{Method}}
& \multirow{2}{*}{\textbf{Backbone}}
& \multicolumn{3}{c|}{\textbf{CIFAR10 (\%)}} 
& \multicolumn{3}{c|}{\textbf{CIFAR100 (\%)}} 
& \multicolumn{3}{c|}{\textbf{ImageNet-100 (\%)}}
& \multicolumn{3}{c|}{\textbf{CUB-200-2011 (\%)}} 
& \multicolumn{3}{c|}{\textbf{Stanford Cars (\%)}} 
& \multicolumn{3}{c|}{\textbf{Oxford Pets (\%)}} 
& \multicolumn{3}{c}{\textbf{Food101 (\%)}} \\
\cmidrule(lr){4-6}
\cmidrule(lr){7-9}
\cmidrule(lr){10-12}
\cmidrule(lr){13-15}
\cmidrule(lr){16-18}
\cmidrule(lr){19-21}
\cmidrule(lr){22-24}
& &  
& All & Old & New
& All & Old & New
& All & Old & New
& All & Old & New
& All & Old & New
& All & Old & New
& All & Old & New \\
\midrule

% =================== GREEDY BLOCK ===================
\multirow{6}{*}{\rotatebox[origin=c]{90}{\textit{Greedy--Hungarian}}}
& SMILE \cite{SMILE} & DINO
& 78.2 & \textbf{99.3} & 67.6
& 61.3 & 70.7 & 42.5
& 39.9 & 87.1 & 16.2
& 41.1 & 67.6 & 27.8
& 33.4 & 58.4 & 21.3
& 54.1 & 66.1 & 47.8
& 34.4 & 64.0 & 19.4 \\
& SMILE \cite{SMILE} & CLIP
& 82.4 & 97.4 & 74.9
& 56.4 & 64.6 & 40.0
& 47.5 & 71.0 & 35.7
& 43.7 & 69.7 & 30.8
& 36.7 & 57.2 & 26.8
& 58.2 & 77.5 & 48.1
& 40.5 & 70.4 & 25.2 \\
& PHE \cite{PHE} & DINO
& 83.0 & \uline{98.0} & 75.5
& 64.8 & 78.8 & 36.9
& 53.1 & 83.5 & 38.1
& 46.9 & 76.0 & 32.4
& 46.3 & \uline{78.3} & 30.8
& 63.3 & 91.3 & 48.6
& \uline{50.0} & \textbf{89.3} & \uline{30.0} \\
& PHE \cite{PHE} & CLIP
& 79.3 & 97.0 & 70.4
& 66.1 & 80.3 & 37.5
& 52.9 & 87.8 & 35.5
& 44.2 & 70.3 & 31.1
& \uline{46.4} & 78.1 & \uline{31.1}
& 64.1 & 86.2 & 52.4
& 47.8 & \uline{88.4} & 27.0 \\
& \textbf{Ours} & DINO
& \cellcolor{ltgray}{\textbf{86.2}} & \cellcolor{ltgray}{95.4} & \cellcolor{ltgray}{\textbf{79.3}}
& \cellcolor{ltgray}{\textbf{72.5}} & \cellcolor{ltgray}{\textbf{85.2}} & \cellcolor{ltgray}{\textbf{47.0}}
& \cellcolor{ltgray}{\textbf{84.1}} & \cellcolor{ltgray}{\uline{94.3}} & \cellcolor{ltgray}{\textbf{63.4}}
& \cellcolor{ltgray}{\uline{52.6}} & \cellcolor{ltgray}{\uline{83.3}} & \cellcolor{ltgray}{\uline{37.2}}
& \cellcolor{ltgray}{42.9} & \cellcolor{ltgray}{78.1} & \cellcolor{ltgray}{25.9}
& \cellcolor{ltgray}{\textbf{81.0}} & \cellcolor{ltgray}{\uline{92.4}} & \cellcolor{ltgray}{\textbf{75.1}}
& \cellcolor{ltgray}{44.5} & \cellcolor{ltgray}{80.4} & \cellcolor{ltgray}{26.2} \\
& \textbf{Ours} & CLIP
& \cellcolor{ltgray}{\uline{84.7}} & \cellcolor{ltgray}{96.0} & \cellcolor{ltgray}{\uline{76.3}}
& \cellcolor{ltgray}{\uline{69.9}} & \cellcolor{ltgray}{\uline{82.8}} & \cellcolor{ltgray}{\uline{44.2}}
& \cellcolor{ltgray}{\uline{83.6}} & \cellcolor{ltgray}{\textbf{96.1}} & \cellcolor{ltgray}{\uline{58.2}}
& \cellcolor{ltgray}{\textbf{58.9}} & \cellcolor{ltgray}{\textbf{87.3}} & \cellcolor{ltgray}{\textbf{44.7}}
& \cellcolor{ltgray}{\textbf{60.4}} & \cellcolor{ltgray}{\textbf{90.6}} & \cellcolor{ltgray}{\textbf{45.8}}
& \cellcolor{ltgray}{\uline{74.9}} & \cellcolor{ltgray}{\textbf{94.6}} & \cellcolor{ltgray}{\uline{64.6}}
& \cellcolor{ltgray}{\textbf{61.2}} & \cellcolor{ltgray}{88.3} & \cellcolor{ltgray}{\textbf{47.3}} \\
\midrule

\multirow{6}{*}{\rotatebox[origin=c]{90}{\textit{Strict--Hungarian}}}
& SMILE \cite{SMILE} & DINO
& 49.9 & \uline{39.9} & 54.9
& 51.6 & 61.6 & 31.7
& 33.8 & 74.2 & 13.5
& 32.2 & 50.9 & 22.9
& 26.2 & 46.6 & 16.3
& 42.9 & 38.7 & 45.1
& 24.2 & 54.3 & 8.8 \\
& SMILE \cite{SMILE} & CLIP
& 51.9 & 19.7 & 68.0
& 46.7 & 55.3 & 29.5
& 35.7 & 41.4 & 32.8
& 34.7 & 55.2 & 24.5
& 32.4 & 46.2 & 25.7
& 40.3 & 37.4 & 41.8
& 33.3 & 56.3 & \uline{21.5} \\
& PHE \cite{PHE} & DINO
& 53.1 & 19.3 & 70.0
& 56.0 & 70.1 & 27.8
& 39.2 & 49.3 & 34.1
& 36.4 & 55.8 & 27.0
& 31.3 & 61.9 & 16.8
& 48.3 & 53.8 & 45.4
& 29.1 & \uline{64.7} & 11.1 \\
& PHE \cite{PHE} & CLIP
& 52.4 & 18.3 & 69.5
& 56.8 & 71.9 & 26.5
& 39.2 & 60.7 & 28.4
& 35.1 & 54.5 & 25.4
& 36.2 & 54.2 & \uline{27.4}
& 52.0 & 52.3 & 51.9
& \uline{33.5} & 58.6 & 20.6 \\
& \textbf{Ours} & DINO
& \cellcolor{ltgray}{\textbf{65.0}} & \cellcolor{ltgray}{\textbf{46.1}} & \cellcolor{ltgray}{\textbf{79.3}}
& \cellcolor{ltgray}{\textbf{64.7}} & \cellcolor{ltgray}{\textbf{77.4}} & \cellcolor{ltgray}{\textbf{39.3}}
& \cellcolor{ltgray}{\textbf{82.6}} & \cellcolor{ltgray}{\uline{92.0}} & \cellcolor{ltgray}{\textbf{63.4}}
& \cellcolor{ltgray}{\uline{43.6}} & \cellcolor{ltgray}{\textbf{63.5}} & \cellcolor{ltgray}{\uline{33.6}}
& \cellcolor{ltgray}{\uline{37.0}} & \cellcolor{ltgray}{\uline{68.1}} & \cellcolor{ltgray}{22.0}
& \cellcolor{ltgray}{\textbf{69.2}} & \cellcolor{ltgray}{\uline{58.5}} & \cellcolor{ltgray}{\textbf{74.8}}
& \cellcolor{ltgray}{30.3} & \cellcolor{ltgray}{60.5} & \cellcolor{ltgray}{15.0} \\
& \textbf{Ours} & CLIP
& \cellcolor{ltgray}{\uline{56.9}} & \cellcolor{ltgray}{31.1} & \cellcolor{ltgray}{\uline{76.3}}
& \cellcolor{ltgray}{\uline{61.6}} & \cellcolor{ltgray}{\uline{75.0}} & \cellcolor{ltgray}{\uline{34.9}}
& \cellcolor{ltgray}{\uline{80.9}} & \cellcolor{ltgray}{\textbf{93.6}} & \cellcolor{ltgray}{\uline{54.9}}
& \cellcolor{ltgray}{\textbf{45.5}} & \cellcolor{ltgray}{\uline{60.7}} & \cellcolor{ltgray}{\textbf{37.8}}
& \cellcolor{ltgray}{\textbf{53.5}} & \cellcolor{ltgray}{\textbf{74.2}} & \cellcolor{ltgray}{\textbf{43.6}}
& \cellcolor{ltgray}{\uline{64.0}} & \cellcolor{ltgray}{\textbf{65.4}} & \cellcolor{ltgray}{\uline{63.3}}
& \cellcolor{ltgray}{\textbf{50.3}} & \cellcolor{ltgray}{\textbf{66.2}} & \cellcolor{ltgray}{\textbf{42.2}} \\
\bottomrule
\end{tabular}
}
\label{tab:acc_app}
\vspace{-1.5em}
\end{table*}

\subsection{The influence of different backbone networks}
To comprehensively evaluate the adaptability and robustness of our approach, we compare it against two representative state-of-the-art methods—SMILE and PHE—under two widely adopted visual backbones, CLIP and DINO,as shown in Tab.~\ref{tab:acc_app}
Specifically, CLIP is a large-scale vision–language contrastive model known for its strong generalization ability, especially in open-world scenarios, while DINO is a self-supervised representation learning method whose distilled features are highly structured and discriminative, offering superior intra-class compactness and inter-class separability.

Across both backbones, our method consistently outperforms SMILE and PHE on all benchmarks. The improvements are observed not only in the overall accuracy (All) but also in the recognition of Old and particularly New categories, where our gains are most prominent. These results demonstrate that our proposed matching strategy remains effective across different feature spaces, showing stronger generalization and robustness compared to existing approaches.

In summary, the experiments with multiple backbones further validate the stability and superiority of our method under diverse representation settings.
\subsection{Computational Complexity Analysis.}

In our framework, the OCD task can be decomposed into three stages: a training stage, a prototype computation stage, and a test stage. Using CLIP as the visual backbone, we measure the running time of each stage and repeat the experiment five times independently. The averaged results and standard deviations are reported in Tab.~\ref{tab:table_time}.All experiments are conducted on a single NVIDIA RTX 3090 GPU with 24 GB memory.

\vspace{-1.25mm}
\begin{table}[H]
% \vspace{-3mm}
\centering
\caption{Three-stage time consumption proportion display table}
\label{tab:table_time}
\resizebox{\linewidth}{!}{
\begin{tabular}{lccc}
\toprule
\textbf{Dataset} & \textbf{Train (ms)} & \textbf{Prototype computation (ms)} & \textbf{Test (ms)} \\
\midrule
CIFAR-10         & 746.506 $\pm$ 1.445 & 667.455 $\pm$ 41.925 & 1496.276 $\pm$ 37.384 \\
CIFAR-100        & 747.667 $\pm$ 3.647 & 734.105 $\pm$ 55.240 & 1504.007 $\pm$ 65.634 \\
ImageNet-100     & 746.663 $\pm$ 1.120 & 666.157 $\pm$ 3.393  & 2324.649 $\pm$ 535.563 \\
CUB-200-2011     & 748.577 $\pm$ 1.020 & 642.528 $\pm$ 1.003  & 1753.667 $\pm$ 44.756 \\
Stanford Cars    & 750.333 $\pm$ 3.086 & 653.369 $\pm$ 1.991  & 1736.116 $\pm$ 64.858 \\
Oxford Pets      & 749.094 $\pm$ 3.262 & 641.062 $\pm$ 0.394  & 1472.476 $\pm$ 72.950 \\
Food101          & 750.889 $\pm$ 0.596 & 647.426 $\pm$ 2.539  & 1724.313 $\pm$ 17.371 \\
\bottomrule
\end{tabular}}
\end{table}

\begin{figure}[H]
    \vspace{-2mm}
    \centering
    \includegraphics[width=1\linewidth]{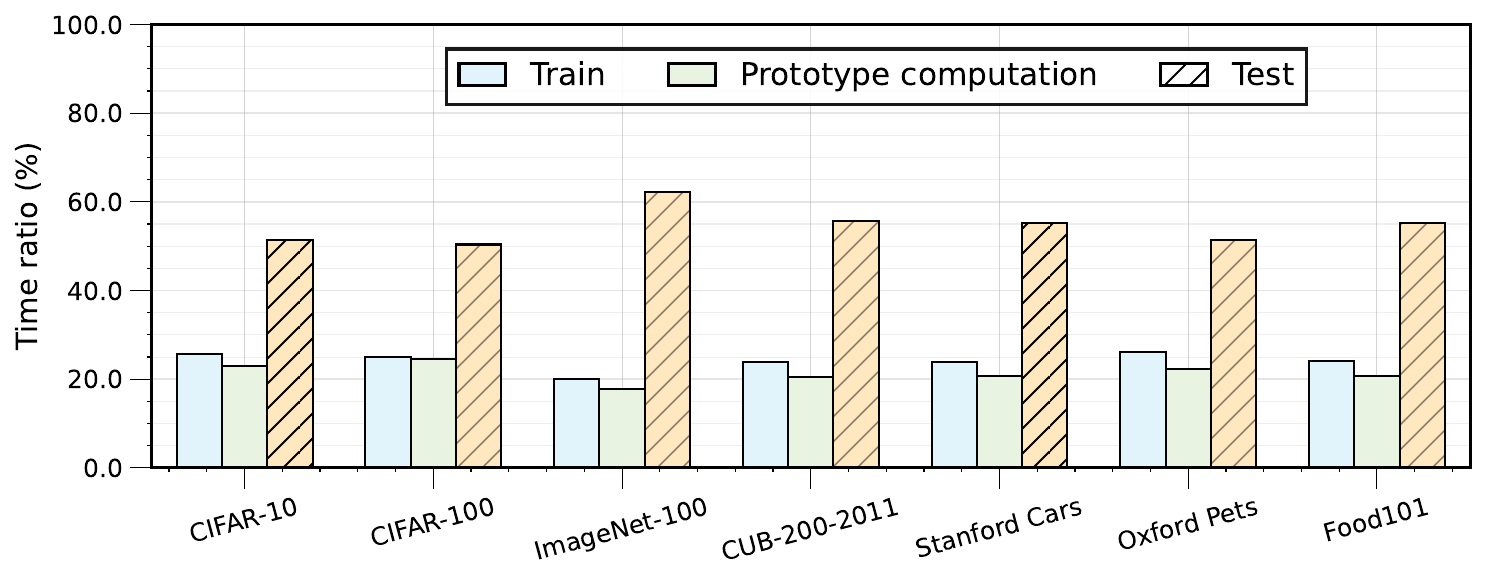}
    \caption{\textbf{Display of the proportion of time consumed in the three stages.}
}
\label{fig:time_ratio}
\end{figure}

\vspace{-3mm}
As shown in the Fig.~\ref{fig:time_ratio}, the time proportion during the test is the longest among the three stages. This is due to the introduction of the TTA algorithm. However, under the premise of improving such high accuracy, we believe it is worthwhile.

\subsection{Training Efficiency Analysis}
% \subsection{Training Efficiency Analysis} 
We provide a comparison of training times between our PHE and the state-of-the-art method, SMILE, with results shown in Table~\ref{tab:training-time-all}. To ensure fairness, all experiments are conducted on an NVIDIA RTX 3090 GPU, with both algorithms trained over 100 epochs using mixed precision. The dataloader parameters are kept consistent across tests, with a batch size of 128. 

According to the results in Table \ref{tab:training-time-all}, our method achieves consistently lower training time across all four datasets compared with both SMILE and PHE. Relative to SMILE, our approach reduces training time by 2752.6 s on CUB, 4023.5 s on SCars, 14843.2 s on Food, and 1572.0 s on Pets. Compared with PHE, our method is also substantially more efficient, yielding reductions of 1847.7 s, 3490.4 s, 10608.2 s, and 1213.4 s on the four datasets, respectively. These results demonstrate that our method provides the fastest training among all compared approaches. This improvement largely stems from our more efficient feature learning strategy, whereas SMILE incurs heavy computational cost due to its supervised contrastive learning that processes two augmented views of each sample, and PHE also involves additional estimation and augmentation overhead.
\begin{table}[ht]
\centering
\caption{Comparison of training times (in second)}
\setlength{\tabcolsep}{10pt}
\label{tab:training-time-all}
\begin{tabular}{l|c|c|c|c}
\hline
Method & CUB & SCars & Food & Pets \\ \hline
SMILE           & 4204.9       & 5846.7        & 27000.7       & 2646.5        \\
PHE      & 3300.0      & 5313.6        & 22765.7        &2287.9         \\ \hline
Ours      & \textbf{1452.3}       & \textbf{1823.2}         & \textbf{12157.5}        &\textbf{1074.5}         \\ \hline
\end{tabular}
\end{table}

\section{Broader Impact and Limitations Discussion}

\subsection{Broader Impact.}
Our method improves on-the-fly category discovery under open-world and streaming settings, which can benefit applications such as long-term perception, large-scale retrieval, and biodiversity monitoring by reducing the need for repeated manual re-labeling. At the same time, automatic discovery of novel categories on uncurated data streams may amplify existing dataset biases or unintentionally cluster sensitive attributes, so any real-world deployment should include careful data curation, monitoring, and human oversight.

\subsection{Limitations.}
Our framework relies on strong pre-trained vision backbones (CLIP and DINOv2) and GPU resources, and its performance may degrade in low-resource scenarios or when such pre-training is unavailable. In addition, we mainly evaluate on standard OCD benchmarks with moderate numbers of novel classes, while more extreme non-stationary streams or very large numbers of emerging categories may increase prototype memory and adaptation instability.

\subsection{A Possible Solution in Future Work.}
Future work could explore more lightweight or distilled backbones for deployment on edge devices, as well as stronger mechanisms for handling long-term distribution shift, e.g., memory-based replay or more robust prototype regularization. Another promising direction is to incorporate multimodal or human feedback to better name, merge, or filter discovered categories, improving both interpretability and safety in practical applications.

% \section{Additional Visualization Analysis}

% \input{sec/rebuttal}
\end{document}